\documentclass[journal]{IEEEtran}
\usepackage{graphicx}
\usepackage{amsmath,amssymb} 

\title{Motion Deblurring for Plenoptic Images}

\author{Paramanand Chandramouli, Paolo Favaro and Daniele Perrone \thanks{The authors are with Institut f\"{u}r Informatik und angewandte Mathematik, University of Bern, Switzerland. E-mail: chandra@iam.unibe.ch, paolo.favaro@iam.unibe.ch, perrone@iam.unibe.ch.}}

\def\ie{\emph{i.e.}}

\begin{document}

\maketitle

\begin{abstract}
We address for the first time the issue of motion blur in light field images captured from plenoptic cameras. We propose a solution to the estimation of a sharp high resolution scene radiance given a blurry light field image, when the motion blur point spread function is unknown, i.e., the so-called blind deconvolution problem. In a plenoptic camera, the spatial sampling in each view is not only decimated but also defocused. Consequently, current blind deconvolution approaches for traditional cameras are not applicable. Due to the complexity of the imaging model, we investigate first the case of uniform (shift-invariant) blur of Lambertian objects, i.e., when objects are sufficiently far away from the camera to be approximately invariant to depth changes and their reflectance does not vary with the viewing direction. We introduce a highly parallelizable model for light field motion blur that is computationally and memory efficient. We then adapt a regularized blind deconvolution approach to our model and demonstrate its performance on both synthetic and real light field data. Our method handles practical issues in real cameras such as radial distortion correction and alignment within an energy minimization framework.
\end{abstract}
 
\section{Introduction}
An important outcome of advances in computational imaging is that of a plenoptic camera which can directly capture the light field information of a scene. In the past few years, plenoptic cameras have entered into the realm of consumer photography \cite{web:lytro,web:ray}. These camera are equipped with capabilities, such as 3D reconstruction and digital refocusing, not possible in traditional devices. This has led to an increased interest in the scientific community in high-quality light field reconstruction. As these commercial cameras are portable, camera shake is sometimes unavoidable and may result in blurry light field images. Similarly, moving objects can also cause the images to appear blurry. 

Until now most research works on light field (LF) imaging have focused on depth estimation and super-resolution. There are no existent algorithms to handle motion blur in LF images. In contrast, motion blur in conventional cameras has been widely studied and current methods achieve remarkable results (see, for instance, \cite{Cho2009,ppr:high_mtn,Xu2010,Levin2011,babacan}). Unfortunately, the imaging mechanism of a conventional camera and LF camera are quite different. Due to the additional microlens array between the main lens and the sensors in the LF camera an LF image consists of a rearranged set of views of the scene that are highly under-sampled and blurred \cite{ng2005light}. Consequently, motion deblurring methods that are applicable for conventional images cannot be adapted in a straightforward manner.

In this paper, we propose a motion deblurring scheme for images captured from a microlens-array based light field camera. An LF image can be related to the high resolution scene texture through a space-variant point spread function (PSF) which models the image formation mechanism. In addition, the sharp scene texture is related to the blurry texture in terms of a motion blur PSF. Modeling the LF image generation by taking into account these effects turns out to be very computationally intensive and memory inefficient. However, we show that, when restricted to a constant depth scenario, it is possible to describe a motion blurred light field image as a linear combination of parallel convolutions (see sec.~\ref{sec:fast}). As a result, the model is extremely computationally and memory efficient. We address the scenario of uniform motion blur. However to handle realistic scenarios, within our framework, we allow for small variations in motion blur across the image. From a single motion blurred LF image, we solve for the high-resolution sharp scene radiance and the motion blur PSF in an energy minimization framework. We demonstrate the performance of our algorithm on real and synthetic images. The quality of results obtained by the proposed method is much higher when compared to outputs of techniques based on conventional deblurring. For real experiments, we use Lytro Illum camera. The images obtained from the LF camera suffer from radial distortion and misalignments. Our method accounts for these effects by adequately modifying the image formation model. Although, we have considered scenes that are significantly far or fronto-parallel planar, our method can be extended to handle scenes with depth variations. However, this would require estimation of depth map and handling of depth discontinuities which we consider to be beyond the scope of this paper.

\section{Related work}
There is no prior work that deals with motion blurred light field reconstruction. However, since our work relates to both light field reconstruction and motion deblurring, we provide a brief overview of some methods developed in both areas. 

\textbf{Single image motion deblurring.}
Motion deblurring involves the joint estimation of a sharp image and a blur kernel. Because of its ill-posedness, motion deblurring is typically addressed by enforcing prior information on the sharp image and on the blur kernel \cite{Cho2009,Fergus2006,Levin2011Understanding,ppr:high_mtn,Xu2010}. Popular choices of prior include Laplace distribution~\cite{Levin2011}, total variation (TV) regularization \cite{Chan1998} or $L_0$ regularization \cite{L0}. Other methods encourage sharp edges by using a shock filter \cite{Cho2009,Xu2010} or a dictionary of sharp edges~\cite{Sun2013}.
For the blur function, the choices for prior include Gaussian distribution~\cite{Xu2010}, or a sparsity-inducing distribution \cite{Fergus2006}.
Given suitable priors on the sharp image and the blur function, motion deblurring is typically solved via a Maximum a Posteriori (MAP) estimation. Levin et al.~\cite{Levin2011Understanding} have shown that a joint MAP estimation of the sharp image and blur function cannot lead to a correct estimate for a wide range of image priors. They show that marginalizing the sharp image and performing a MAP estimation on the blur function alone can correctly estimate the blur kernel. The marginalization of the sharp image is however computationally challenging, and therefore various approximations are used in practice~\cite{Fergus2006,Levin2011,babacan}. 
Despite the theoretical analysis of Levin et al.~\cite{Levin2011Understanding}, many methods successfully use a joint MAP estimation and achieve state of the art results~\cite{Cho2009,ppr:high_mtn,Xu2010,logtv}. Recently, Perrone and Favaro \cite{Perrone2014} have clarified this apparent inconsistency. They confirmed the results of Levin et al.~\cite{Levin2011Understanding} and showed that particular implementation details make many methods in the literature work and that they do not in practice solve a joint MAP estimation.

The aforementioned deblurring methods assume uniform blur across the image plane. Techniques exist that address non-uniform blur due to rotational camera motion by introducing addition dimensions in the motion blur PSF \cite{ppr:gupta_eccv,ppr:hu_bmvc,ppr:mbrown,ppr:whyte}. Hirsch et al. propose an efficient implementation scheme wherein non-uniform blur can be modeled through piece-wise uniform blurs while still retaining the global camera motion constraint \cite{ppr:hirsch_iccv}. Nonetheless, it has been observed in \cite{Kohler2012} that in realistic scenarios, the performance of algorithms that explicitly handle rotational motion are not necessarily better than that of the methods that assume uniform blur. Motion blur can also vary within an image because of parallax in 3D scenes. Techniques proposed in \cite{Sorel2008,XuICCP} address this problem when there are two observations available and the camera motion is restricted to in-plane translations. While in \cite{Param2013}, a deblurring scheme which addresses non-uniform blur for bilayer scenes using two blurred images is proposed, Hu et al. address the problem with a single observation \cite{HuDepth2014}. Other methods attempt to solve general space-varying motion deblurring by estimating locally uniform blur and carefully interpolate them to cope with regions having poor texture~\cite{Ji2012}, or by iteratively employing uniform deblurring and segmentation algorithms~\cite{Hyun2013}.

\textbf{Light field capture}
The basic lenslet-based plenoptic camera was first developed by Adelson and Wang for inferring scene depth from a single snapshot \cite{ppr:adelson}. The portable design by Ng et al. with the use of microlens arrays triggered the development of handheld light field cameras \cite{ng2005light}. The main drawback of image reconstruction in this camera was its limited spatial resolution. To overcome this problem, Lumsdaine and Georgiev designed the focused plenoptic camera wherein the microlenses were focused at the image plane of the main lens, thereby enabling rendering at higher resolution for certain depths \cite{GeorgievSR}. Perwa{\ss} and Wietzke proposed a further modification in the design which included microlens arrays with microlenses having different focal lengths \cite{Perwass2012}. Recently, Wei et al. proposed to introduce irregularity in the lens design to achieve better sampling of light fields \cite{Wei2015}. Light field cameras have also been built by using masks instead of microlens arrays \cite{Ashok2007,Marwah2013}.  Another approach to capture light field is by using camera arrays or a gantry-based moving camera \cite{levoy1996light,vaish2004using}. In this paper, we restrict our attention to microlens array-based light field cameras.

\textbf{Calibration, depth estimation and super-resolution} 
Since capturing light fields from plenoptic cameras suffers from undersampling in the spatial domain, various algorithms have been proposed to super-resolve scene radiance. In \cite{GeorgievSR}, the authors propose to render high resolution texture directly from light field data by projecting the LF image onto a fine grid of scene radiance. In \cite{Bishop}, Bishop and Favaro model the image formation through geometric optics and propose a PSF-based model to relate the LF image and high resolution scene radiance along with the camera parameters and scene depth. They propose a two-step procedure to estimate the scene depth map using view correspondences and thence the high resolution scene radiance. Wanner and Goldl\"{u}cke propose a technique to accurately estimate disparity using epipolar image representation and then super-resolve 4D light fields in both spatial and angular directions \cite{ppr:Goldluecke13}. Broxton et al. propose a 3D-deconvolution method to reconstruct a high resolution 3D volume using a PSF-based model for the scenario of light field microscopy \cite{Broxton:13}. We also found out that this work suggests a fast computational scheme similar to the one proposed here (a very short explanation is given in a paragraph in sec.3.4 of \cite{Broxton:13}). However, our and this scheme were developed simultaneously. Moreover, our scheme includes the case of motion blur.

Recently, light field processing algorithms that also address practical issues in handling data from consumer cameras have been proposed. Danserau et al. propose a decoding, calibration and rectifying procedure for lenselet-based plenoptic cameras \cite{ppr:calib}. They develop a parameterized plenoptic camera model that relates pixels to rays in 3D space. Bok et al. propose another calibration scheme using line features \cite{BokEccv2014}. Cho et al. \cite{ppr:calib_tai} develop a method for rectification and decoding of light field data. They also develop a scheme for rendering texture at a higher resolution using a learning-based interpolation method. Xu et al. propose a decoding scheme which does not need a calibration image, instead the model parameters are estimated in a optimization scheme \cite{XuACCV}. They also propose a 4D demosaicing scheme based on kernel regression. Huang and Cossairt propose a dictionary learning based color demosaicing scheme for plenoptic cameras \cite{Xiangcvprw}. In \cite{YuCVPR12}, Yu et al. propose to perform demosaicing while rendering the refocused image. Fiss et al. develop a scheme to refocus plenoptic images using depth adaptive splatting \cite{FissICCP2014}. In \cite{Mitra2014}, a hybrid imaging system with a conventional and a plenoptic camera is proposed to achieve super-resolution.

Techniques also exist that focus on obtaining accurate depth maps from plenoptic images. For example, in \cite{TaoICCV13}, Tao et al. combine both the correspondence and defocus cues to estimate scene depth. Motivated by the idea of active wavefront sampling, Heber et al. develop a scheme to arrive at a high quality depth map \cite{HeberRP13}. Sabater et al. propose to estimate disparity from mosaiced view correspondences instead of demosaiced views to achieve more accuracy \cite{Sabater}. Finally, Liang and Ramamoorthi  provide an interesting analysis of the plenoptic camera using light field transport approach and propose a general framework to model all the components of the image formation process \cite{Liang}. 

\textit{Contributions.} The contributions of our work can be summarized as: i) We introduce the problem of motion deblurring of light field images from a plenoptic camera and provide the first solution to it. ii) We propose a computationally and memory efficient imaging model for motion blurred LF images. iii) We solve a joint blind deconvolution and super resolution problem. iv) We handle radial distortion correction and alignment within our energy minimization framework.

\section{Imaging model}
\label{sec:model}

In this section we introduce notation and the image formation model for a motion blurred light field image. The model relates the light field image captured by a plenoptic camera with the scene radiance via an explicit PSF. We describe the imaging model assuming uniform motion blur. Following the convention in \cite{Bishop}, we consider the microlens array plane to be the domain on which the high-resolution scene radiance $f$ is defined. The light field image $l$ is defined on the image sensor plane. Let $\mathbf{p}$ denote the coordinates of a point on the microlens array and $\mathbf{x}$ denote a pixel location on the sensor plane. We have $\mathbf{p}=[p_1~p_2]^T$, and $\mathbf{x}=[x_1~ x_2]^T$, where $p_1,p_2,x_1,x_2$ are integers. A pixel of the LF image $l(\mathbf{x})$ is related to texture elements $f(\mathbf{p})$ through a space-varying point spread function $h(\mathbf{x},\mathbf{p})$ as 
 \begin{align}
 l(\mathbf{x})=\sum_{\mathbf{p}}h(\mathbf{x},\mathbf{p})f(\mathbf{p}).
 \label{eqn:lfgen}
 \end{align}
In general, the PSF $h$ depends on the scene depth map and the parameters of the plenoptic camera. An explicit formula is available in \cite{Broxton:13,Bishop}.  
A relative motion between the camera and the scene during the exposure interval causes the light field image to be motion blurred. The observed motion blurred LF image can be written as a weighted sum of multiple light field images corresponding to shifted scene texture. The weights define the motion blur PSF $h_m$. Since the depth is constant, we can express the motion blurred light field $l_m$ as 
\begin{align}
l_m(\mathbf{x})=\sum_{\mathbf{p}}h(\mathbf{x},\mathbf{p})\sum_{\mathbf{q}}h_m(\mathbf{q})f(\mathbf{p}-\mathbf{q}) 
=\sum_{\mathbf{p}}h(\mathbf{x},\mathbf{p})g(\mathbf{p})
 \label{eqn:lfmb1}
\end{align}
where $g(\mathbf{p})\doteq \sum_{\mathbf{q}}h_m(\mathbf{q})f(\mathbf{p}-\mathbf{q}).$
\subsection{Convolution-based LF image generation}
\label{sec:fast}

A direct implementation of eq.~(\ref{eqn:lfgen}) is practically not feasible due to the memory and computation requirements. For instance, if the scene radiance and the light field image were of the order of mega pixels, the number of elements necessary to store $h$ would be of the order of $10^{12}$. Although $h$ is sparse, performing the sum and product calculation in eq.~\eqref{eqn:lfgen} would still be computationally intensive. We address these shortcomings by exploiting the structure of LF PSFs. As a result we obtain an exact and efficient implementation of LF image generation that is highly parallelizable. 

\begin{figure*}[t!]
\centering
\includegraphics[width=\textwidth]{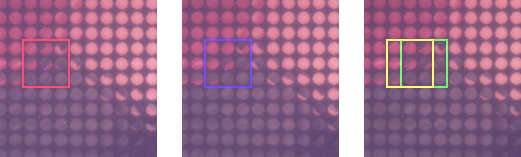}\\ \ \\
\includegraphics[width=\textwidth]{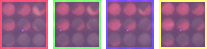}
\caption{Illustration of the periodicity of a LF image. Row 1: three different LF images captured by horizontally shifting texture. Zoomed-in patches whose colors indicate the portion that was cropped from the images in the first row.
\label{fig:PSF_shift}}
\end{figure*}
We first provide an intuitive explanation of the key idea that we exploit. Then, we present it formally. Although we consider a rectangular arrangement of microlenses, the idea holds even for hexagonal arrangement as seen in section \ref{sec:hex}. 
Consider two LF images that are generated with the same texture, but differ by a small shift along the $X$ axis. If the shift in the texture $f$ matches the distance between two microlenses, the two LF images will turn out to be exact copies of each another up to a shift equal to the number of pixels under a microlens. To illustrate this fact, we captured a few real light field images by shifting a poster from left to right while keeping our LF camera still. In the first row of Fig.~\ref{fig:PSF_shift}, we show the same region cropped from three light field images. We can observe the shift of the edge pattern in these images as the poster is shifted. Now compare the first with the third image. They are (almost) identical up to a shift of $1$ microlens. In the second row of Fig.~\ref{fig:PSF_shift}, we show zoomed-in patches for better illustration. While the red and green patches are almost similar, the yellow patch has a shift corresponding to one microlens (with respect to the red patch). The blue patch corresponds to a texture with an intermediate shift. Thus, we can see that there is a periodic structure in the light field. Note that there is an artifact in the central microlens of the red patch. This is not due to the scene texture but because of a damaged sensor pixel.

To formalize this periodicity, we need to introduce some basic notation. Suppose there are $J{\times}J$ pixels under each microlens in the sensor plane. A pixel location $\mathbf{x}$ can be written as $\mathbf{x}=\mathbf{k}J+\mathbf{j}$, where $\mathbf{j}=[j_1~j_2]$, $j_1,j_2\in[0,\dots,J-1]$, and $\mathbf{k}=[k_1~k_2]$ with $k_1,k_2\in Z$. It should be noted that while all those pixels with the same $\mathbf{j}$ correspond to a view or a sub-aperture image (like one image from a camera array), those with the same $\mathbf{k}$ correspond to the image within a single microlens. We decompose the coordinates $\mathbf{p}$ on the microlens plane in a similar fashion. If two microlenses are $D$ units apart, then we write $\mathbf{p}=\mathbf{b}D+\mathbf{t}$, where $\mathbf{t}=[t_1~t_2]$, $t_1,t_2\in[0,\dots,D-1]$, and $\mathbf{b}=[b_1~b_2]$, $b_1,b_2\in Z$. Note that $D$ defines the resolution at which the scene radiance is defined. A larger value of $D$ defines radiance on a finer grid while a smaller value would lead to a coarse resolution.

Now consider two points $\mathbf{p}$ and $\mathbf{q}$ located in different microlenses in such a way that their relative position with respect to the microlens center is the same (\ie, both have the same value of $\mathbf{t}$). Then, the PSFs $h(\mathbf{x},\mathbf{p})$ and $h(\mathbf{x},\mathbf{q})$ will be shifted versions of each another. Technically, this is because the PSF of the light field image is given by the intersections of the blur discs of the main lens and the microlens array \cite{Bishop}. If there is a shift in the position of the point light source which exactly corresponds to the microlens diameter, then the resulting intersection pattern will be exactly identical to the original intersection pattern but with a shift because of the regularity of the microlens array. Thus, we can write
\begin{align}
h\left(\mathbf{x},\mathbf{p}\right) = h\left(\mathbf{x}-J\mathbf{y},\mathbf{p}-D\mathbf{y}\right)
\label{eqn:PSFshift}
\end{align}
for any $\mathbf{y}$, where $\mathbf{y}=[y_1~y_2]$, $y_1,y_2\in Z$. The first observation is that we only need to store $J^2\times D^2$ blur kernels, as the others can be obtained by using eq.~\eqref{eqn:PSFshift}. Secondly, the spatial spread of the PSF will be limited because of the limited extent of main lens blur disc. Thus, we have obtained a memory efficient representation of the PSF $h$. By replacing $\mathbf{p}$ by $\mathbf{b}D+\mathbf{t}$ in eq.~(\ref{eqn:lfmb1}), and using eq.~(\ref{eqn:PSFshift}) we get
\begin{align}
l_m(\mathbf{x})=\sum_{\mathbf{t}}\sum_{\mathbf{b}}h(\mathbf{x}-\mathbf{b}J,\mathbf{t})g(\mathbf{b}D+\mathbf{t})
\end{align}
Then, by expressing $\mathbf{x}$ as $\mathbf{x}=\mathbf{k}J+\mathbf{j}$, we get
\begin{align}
l_m(\mathbf{k}J+\mathbf{j})=\sum_{\mathbf{t}}\sum_{\mathbf{b}}h(\mathbf{k}J+\mathbf{j}-\mathbf{b}J,\mathbf{t})g(\mathbf{b}D+\mathbf{t})
\end{align}
Let $\hat{g}(\mathbf{b},\mathbf{t})\doteq g(\mathbf{b}D+\mathbf{t})$, $\hat{l}_m(\mathbf{k},\mathbf{j})\doteq l_m(\mathbf{k}J+\mathbf{j})$ and $\hat{h}_\mathbf{j}(\mathbf{k},\mathbf{t})\doteq h\left(\mathbf{k}J+\mathbf{j},\mathbf{t}\right)$ (\ie, just a rearrangement). Then, for every value of $\mathbf{j}$, we have
\begin{align}
\hat{l}_m(\mathbf{k},\mathbf{j})=\sum_{\mathbf{t}}\sum_{\mathbf{b}}h\left((\mathbf{k}-\mathbf{b})J+\mathbf{j},\mathbf{t}\right)\hat{g}(\mathbf{b},\mathbf{t})
\label{eqn:lfconv}
\end{align}
and finally
\begin{align}
\nonumber
\hat{l}_m(\mathbf{k},\mathbf{j})=&\sum_{\mathbf{t}}\sum_{\mathbf{b}}\hat{h}_\mathbf{j}\left(\mathbf{k}-\mathbf{b},\mathbf{t}\right)\hat{g}(\mathbf{b},\mathbf{t})
\ \\
=&\sum_{\mathbf{t}}\left( \hat{h}_\mathbf{j}\left(\cdot,\mathbf{t}\right) \ast \hat{g}\left(\cdot,\mathbf{t}\right)\right)(\mathbf{k})
\label{eqn:lfconvfinal}
\end{align}
where $\ast$ denotes the convolution operation. Eq.~(\ref{eqn:lfconv}) indicates that we can arrive at the LF image by performing convolutions for every possible value of $\mathbf{t}$ and $\mathbf{j}$. Note that these convolutions are completely independent and therefore can be executed in parallel. Also, we need only $J^2\times D^2$ PSFs that are denoted by $\hat{h}_\mathbf{j}\left(\cdot,\mathbf{t}\right)$ for LF image generation.


%

\subsection{Hexagonal arrangement}
\label{sec:hex}
\begin{figure}
\begin{center}
\begin{tabular}{c}
\includegraphics[trim=10 40 0 40,clip,width=250pt]{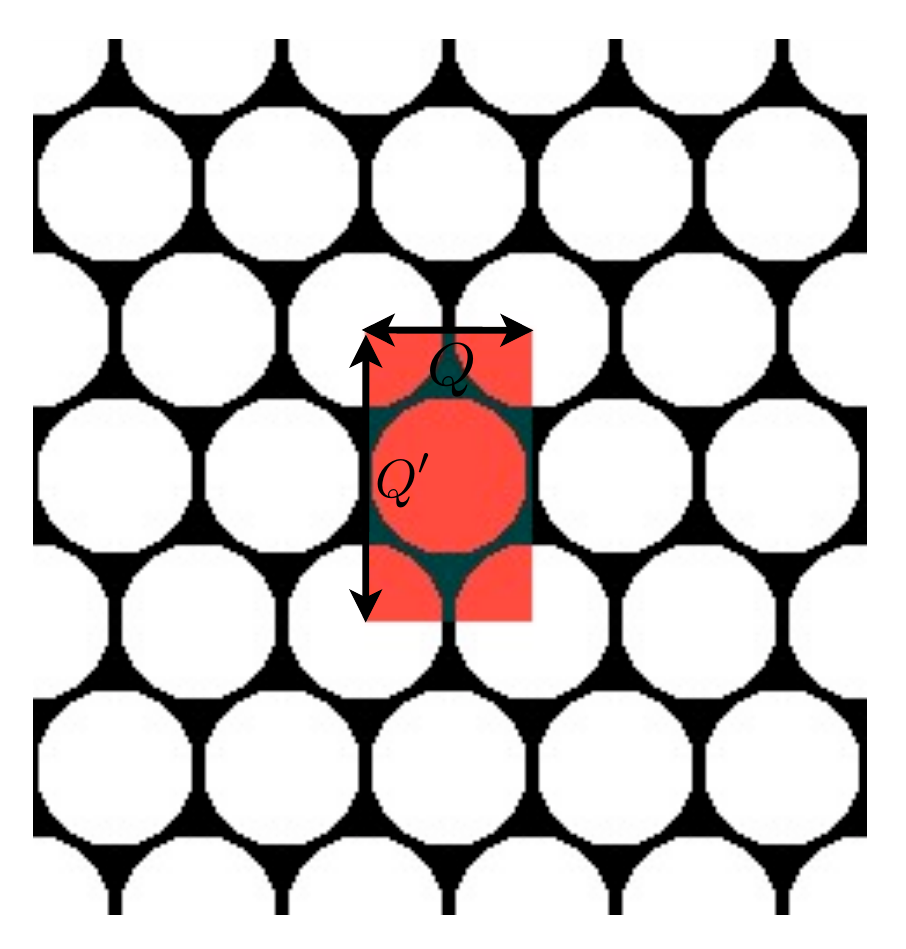} \\
\end{tabular}
\caption{Hexagonal arrangement of microlenses.\label{fig:hexReg}}
\end{center}
\end{figure}

In consumer plenoptic cameras, the microlenses are arranged on a hexagonal grid. Fig. \ref{fig:hex} (a) shows an ideal hexagonal arrangement of microlens arrays. The basis for our convolution-based generation model was that the intersection pattern of the main lens blur disc and the microlens blur disc repeats periodically as we traverse along the microlens array plane. For the hexagonal arrangement, if we consider the set of pixels marked in red color in Fig. \ref{fig:hexReg} (a) as one block and traverse in steps of this block, then the periodicity property holds. While in section \ref{sec:fast}, for the rectangular arrangement, one block would correspond to one microlens with $J{\times}J$ pixels, in the hexagonal case, the size of the block is $Q{\times}Q^{\prime}$ pixels as indicated in Fig. \ref{fig:hexReg} (a). Analogously, for defining the scene texture, we consider that on the microlens array plane, there are $B{\times}B^{\prime}$ units per block as against $D{\times}D$ units in the rectangular arrangement. i.e., we consider that the view index $\mathbf{j}=[j_1~j_2]$, $j_1\in[0,\dots,Q^{\prime}-1]$, and $j_2\in[0,\dots,Q-1]$, and similarly, $\mathbf{t}=[t_1~t_2]$, $t_1\in[0,\dots,B^{\prime}-1]$, and $t_2\in[0,\dots,B-1]$. With these changes, based on our discussion in section \ref{sec:fast}, one can see that the LF image generation model in eq. (\ref{eqn:lfgen}) can be implemented using the convolution-based approach given in eq. (\ref{eqn:lfconvfinal}) even for the hexagonal arrangement. In this scenario, we would need $Q{\times}Q^{\prime}{\times}B{\times}B^{\prime}$ different LF PSFs which are denoted by $\hat{h}_\mathbf{j}\left(\cdot,\mathbf{t}\right)$.

\section{Plenoptic Motion Deblurring}
In this section, we describe our blind deconvolution algorithm wherein we solve for the sharp scene radiance from a blurry light field image. Initially we assume ideal imaging conditions and uniform motion blur, and describe our technique. We subsequently discuss the modifications in our algorithm to account for small variations in motion blur and practical effects such as radial distortion, vignetting, and misalignment.

\subsection{LF blind deconvolution}
\label{sec:lfbd}
 We consider that the knowledge of scene depth value and the camera parameters are available and based on these values, we calculate the LF PSF using the closed-form expression given in \cite{Bishop}. Then, to solve for both the motion blur PSF and the latent sharp texture from a single blurred LF image, we jointly estimate the motion blur PSF and latent texture by using an alternating minimization procedure. 

Based on the model in eq.~\eqref{eqn:lfconvfinal} of section~\ref{sec:model}, the objective function can be written as 
\begin{align}
\nonumber
\min_{f,h_m} \displaystyle \frac{1}{2} \left\|\sum_{\mathbf{p}}h(\mathbf{x},\mathbf{p})\sum_{\mathbf{q}}h_m(\mathbf{q})f(\mathbf{p}-\mathbf{q}) - l_m\right\|_2^2 + \lambda\|f\|_{BV}\\
\mbox{subject to }  h_m \succcurlyeq 0,\quad \|h_m\|_1 = 1
\label{eq:onestep}
\end{align}
where $\|f\|_{BV} \doteq \int ||\nabla f(\mathbf{p})||_2 d\mathbf{p}$,  with $\nabla f \doteq [f_x~f_y]^T$, is the total variation of $f$, and $\lambda>0$ regulates the amount of total variation. The norm $\|\cdot\|_1$ corresponds to the $L^1$ norm.

To solve the above problem, we follow an approach similar to the projected alternating minimization algorithm of \cite{Perrone2014}. The optimization is solved via an iterative procedure based on gradient descent, where at each iteration $n$ the current estimate of the sharp texture is given by
\begin{equation}
\begin{array}{ll}
\bar{g} &\hspace{-3mm}\doteq h^{n-1}_m \ast f^{n-1}\\
\bar{l}_m(\mathbf{x}) &\hspace{-3mm}\displaystyle \doteq \sum_{\mathbf{p}} h(\mathbf{x},\mathbf{p})\bar{g}(\mathbf{p})\\
\bar{g}^{\ast}(\mathbf{q}) &\hspace{-3mm}\displaystyle \doteq \sum_{\mathbf{x}} h(\mathbf{x},\mathbf{q})(\bar{l}_m(\mathbf{x})-l_m(\mathbf{x}))\\
f^{n} &\hspace{-3mm}= \displaystyle f^{n-1} -\! \epsilon \left[\overline{h}^{n-1}_m \ast \bar{g}^{\ast}  - \lambda \nabla \cdot \frac{\nabla f^{n-1}}{\| f^{n-1}\|_{BV}}\!\right]
\label{eqn:fupdate}
\end{array}
\end{equation}
for some step $\epsilon>0$ and where $\overline{h}_m(\mathbf{q}) = h_m(-\mathbf{q})$. For the blur kernel $h_m$ the iteration is
\begin{equation}
\begin{array}{ll}
\check{g} &\hspace{-3mm}\doteq h^{n-1}_m \ast f^{n}\\
\check{l}_m(\mathbf{x}) & \hspace{-3mm}\displaystyle \doteq \sum_{\mathbf{p}} h(\mathbf{x},\mathbf{p})\check{g}(\mathbf{p})\\
\check{g}^{\ast}(\mathbf{q}) &\hspace{-3mm}\displaystyle \doteq \sum_{\mathbf{x}} h(\mathbf{x},\mathbf{q})(\check{l}_m(\mathbf{x})-l_m(\mathbf{x}))\\
\hat h_m^{n-2/3} &\hspace{-3mm} = h_m^{n-1} - \epsilon \left[\overline{f}^{n} \ast  \check{g}^{\ast} \right]
\label{eqn:hupdate}
\end{array}
\end{equation}
The last updated $\hat h_m^{n-2/3}$ is used to set $h_m^{n}$ by using the following sequential projections
\begin{align}\label{eq:stepsh2}
h_m^{n-1/3} \leftarrow \max\{\hat h_m^{n-2/3},0\},\quad~~
h_m^{n} \leftarrow  \frac{h_m^{n-1/3}}{\| h_m^{n-1/3}\|_1}
\end{align}
Notice that all steps can be computed very efficiently via convolutions as presented in sec.~\ref{sec:fast}.

\subsection{Practical implementation}
\label{sec:practical} 
In this subsection, we use matrix vector notation in our description. For convenience, we use the same notation $f$ to denote a vectorized version of the sharp image, and $l_m$ to denote the vectorized blurry light field image. The matrices $H$ and $H_m$ denote the light field and motion blur PSFs, respectively. The model in eq. (\ref{eqn:lfmb1}) can then be re-written as $l_m = HH_mf$. However, as discussed subsequently, we modify our observation model to relate to the real world data.

\begin{figure}
\begin{center}
\begin{tabular}{cc}
\includegraphics[trim=0 0 0 0,clip,width=115pt,height=90pt]{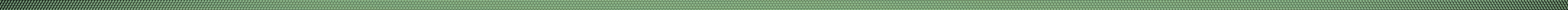} &
\includegraphics[trim=10 10 10 10,clip,width=115pt,height=90pt ]{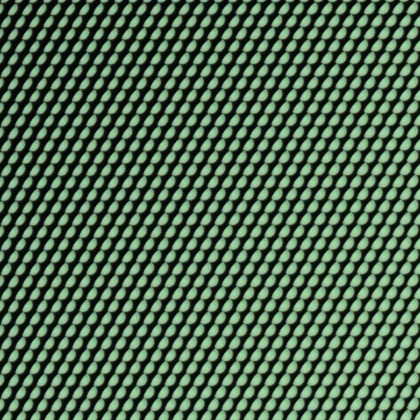} \\
(a) & (b)\\
\end{tabular}
\caption{Cropped white plenoptic image illustrating (a) misalignments, and (b) vignetting.\label{fig:hex}}
\end{center}
\end{figure}

In real cameras, the microlens array is not necessarily aligned with respect to the CCD sensor plane. This effect is apparent in Fig. \ref{fig:hex} (a) which shows a crop of size $50{\times}7728$ pixels from a white image captured with Lytro Illum camera. In Fig. \ref{fig:hex} (a), the gap between adjacent rows of microlenses is tilted instead of being horizontal. Furthermore, in real plenoptic images, the distance between adjacent microlens centers in terms of pixels need not be an integer. Hence, we need to apply an affine mapping which would align the raw image to a regular grid model as shown in Fig. \ref{fig:hexReg} with both $Q$ and $Q^{\prime}$ as integers. Let $l_m$ denote the raw image of a blurry light field captured from a plenoptic camera. We consider that a warping matrix denoted by $W$ when applied on LF image generated according from our model $HH_mf$ aligns it with the raw image $l_m$. The affine transformation remains fixed for a camera and typically its parameters are stored as metadata in plenoptic images \cite{web:toolb}. 


\begin{figure*}[htb!]
\begin{center}
\begin{tabular}{ccc}
\includegraphics[width=164pt,height=125pt]{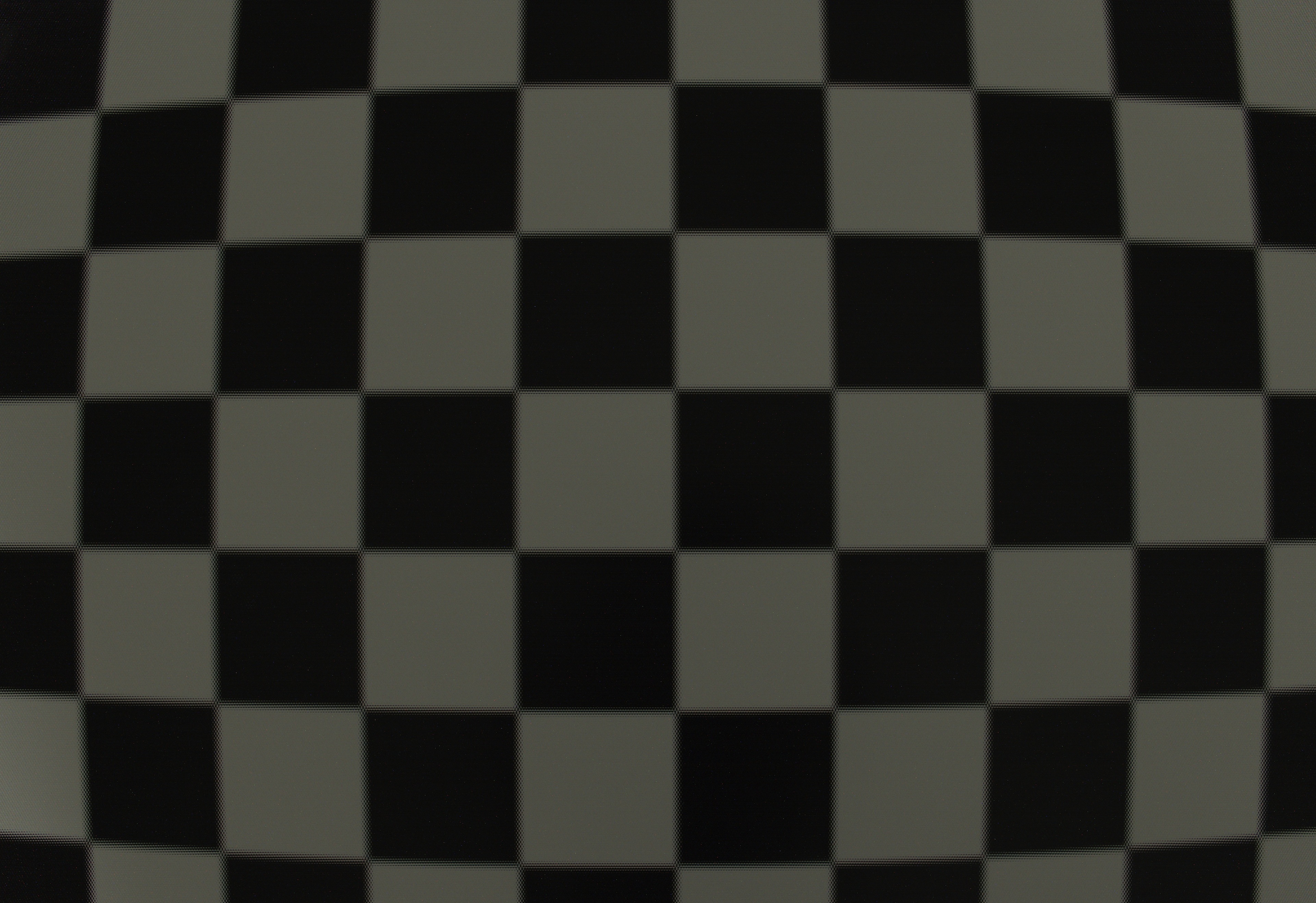}&\hspace{-12pt}
\includegraphics[width=164pt,height=125pt]{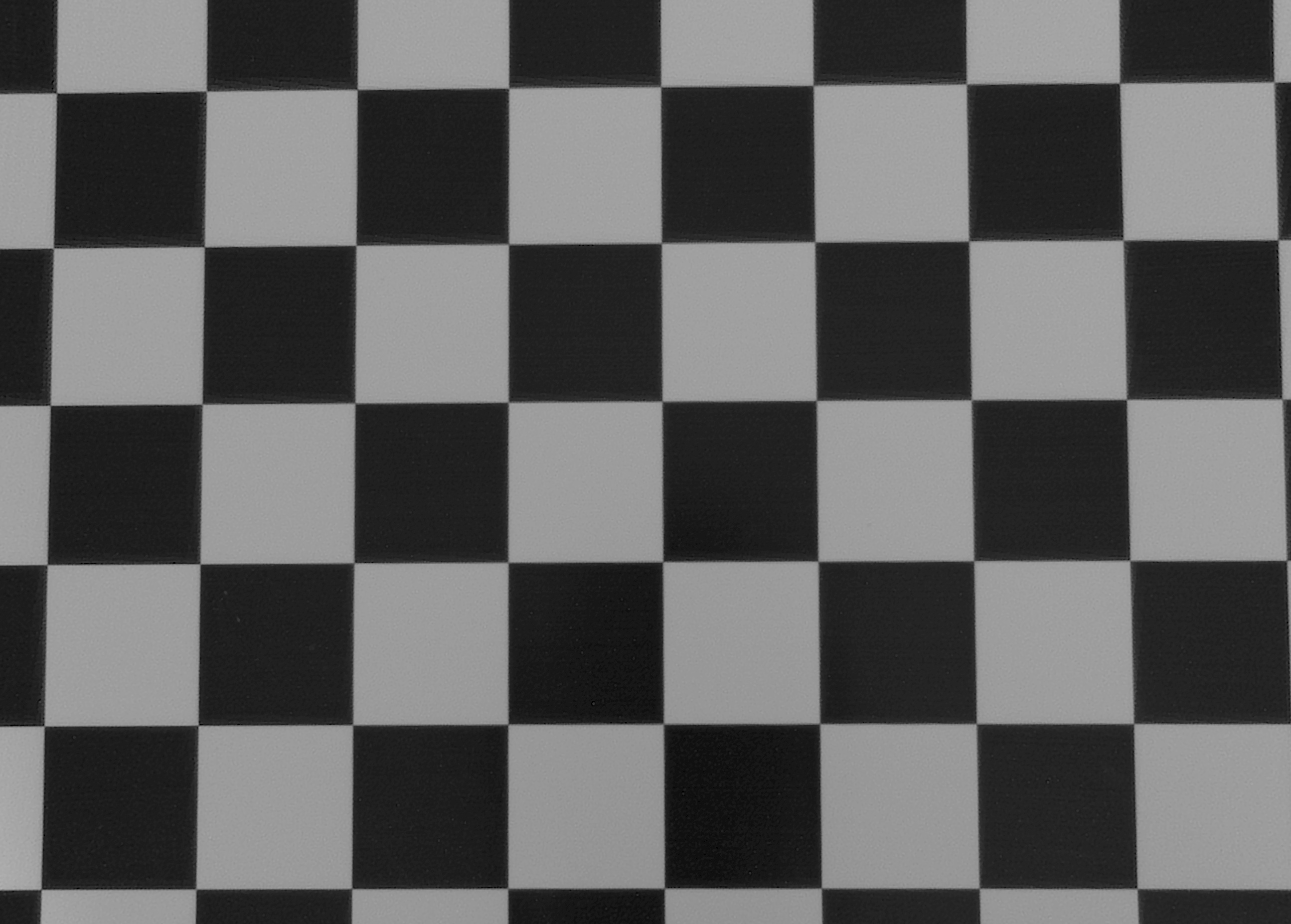}&\hspace{-12pt}
\includegraphics[width=164pt,height=125pt]{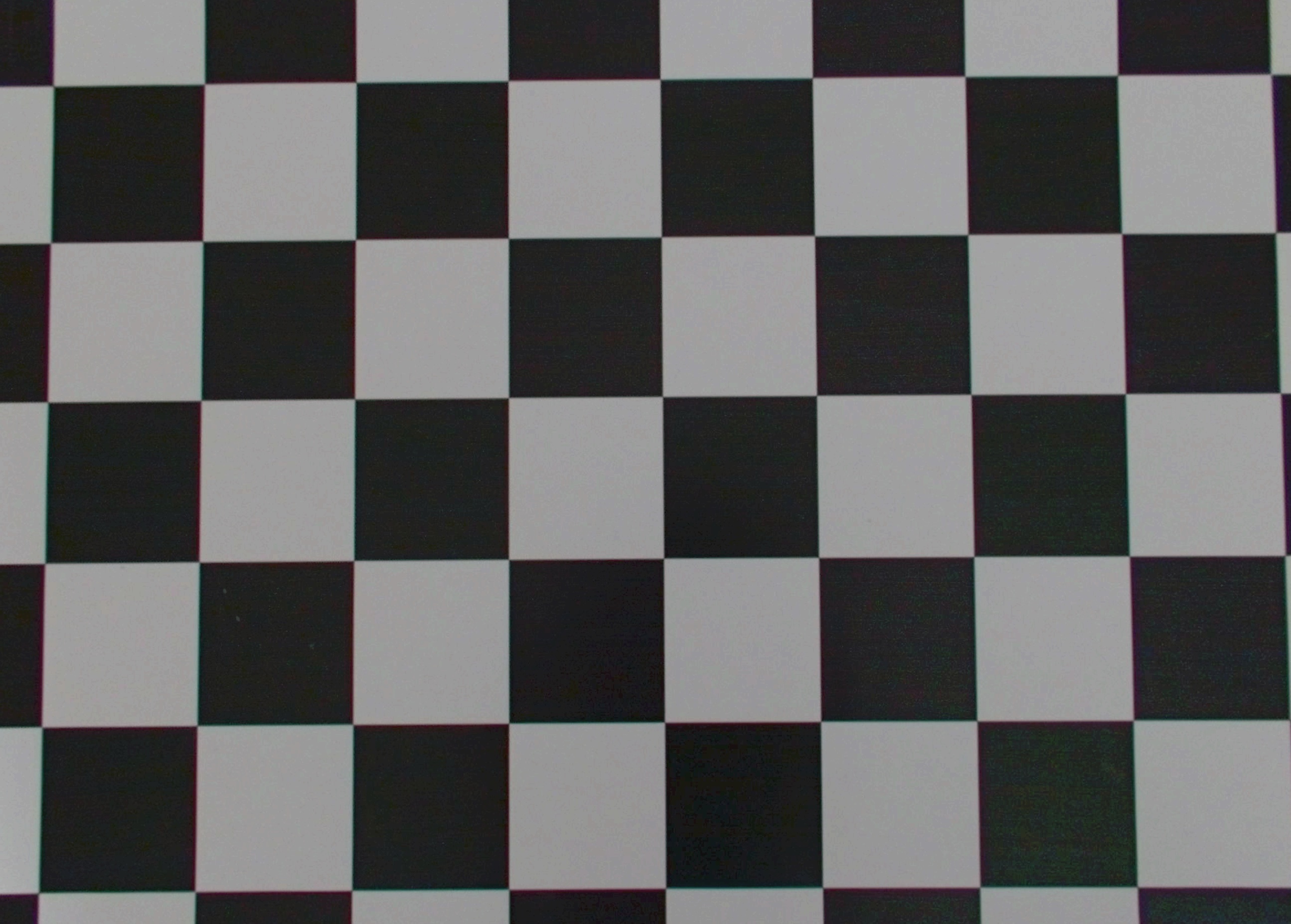}\\
\footnotesize{(a)}&\footnotesize{(b)}&\footnotesize{(c)}
\end{tabular}
\end{center}
\caption{Radial distortion correction. (a) Raw LF image. (b) Estimated texture with distortion correction. (c) Lytro rendering.
\label{fig:rect}}
\end{figure*}

In microlenses that are close to the corners of the plenoptic image, we observe optical vignetting wherein a set of pixels under a microlens do not receive sufficient light from the scene. This is because light rays entering at extreme angles get blocked. Fig. \ref{fig:hex} (b) shows an example of the vignetting effect on LF images. It contains a region cropped at the top left corner of a white plenoptic image. The image pixels under the influence of vignetting do not follow our image formation model. Hence, to avoid using the affected pixels, we use a mask $M$ generated by thresholding a white image. The entries of the mask $M$ will be zero at pixels that are affected by vignetting and one at other locations. To compensate for the angular sensitivity, we apply white image normalization on the captured plenoptic images. In raw images of a camera, we also find ``hot pixels'' due to sensor defects. The locations of the ``hot pixels'' can be determined by capturing a completely dark scene from the camera. These pixel locations are also set to zero in our mask $M$. We also mask the edge pixels of a microlens since they suffer from noise due to demosaicing \cite{ppr:calib}. To summarize, the mask $M$ blocks the pixels affected by vignetting, ``hot pixels'', and pixels present at the border of a microlens.

Yet another effect in real camera is that of radial distortion. For the LF camera which we use in our experiments, we observed significant extent of barrel distortion. Consequently, in our model, we express the LF image as
\begin{align}
l_m = HRH_mf = HRg
\end{align}
where $R$ is a transformation which radially distorts the motion blurred scene texture $g$.
Since we model radial distortion to occur on the scene texture, we can use calibration techniques of conventional cameras such as \cite{web:calib} to determine the distortion parameters. 

In our experiments, we initially determined the radial distortion parameters using the toolbox in \cite{web:calib}. A set of images of a checkerboard pattern was captured from different viewpoints. From each LF image, we estimated the scene texture by ignoring the radial distortion. These images were used as input to arrive at the distortion parameters. Fig. \ref{fig:rect} shows an example wherein radial distortion correction was applied on a checkerboard plenoptic image. The raw image denoted by $l$ is shown in Fig. \ref{fig:rect} (a). Since motion blur is not involved, we obtain the texture by minimizing the error term $\frac{1}{2}\left\|M{\odot}\left(WHRf-l\right)\right\|_2^2$ along with regularization to arrive at the image shown in Fig. \ref{fig:rect} (b). The rendering obtained by Lytro software of the same image (Fig. \ref{fig:rect} (c)) is close to the image in Fig. \ref{fig:rect} (b).
 
By considering the effect of affine warping, masking and radial distortion, the data term of our objective function can be written as 
\begin{align}
E_{\mbox{data}}=\frac{1}{2}\left\|M{\odot}\left(WHRH_mf-l_m\right)\right\|_2^2
\end{align}
where $\odot$ denotes the Hadamard product (point-wise multiplication). Similarly, the data cost can expressed in terms of the motion PSF $h_m$ by consider it to be a vector and the sharp image as a matrix denoted by $F$ .

In order to deal with small variations in motion blur, we modify our algorithm by dividing the image into a few overlapping patches and consider that the motion blur is constant within a patch \cite{ppr:hirsch_iccv}. We also enforce a similarity constraint for the motion blur kernels across patches. Excluding the TV prior on the sharp texture, the cost in our formulation in terms of motion blur PSFs can then be expressed as 
\begin{align}
\nonumber
E_{\mbox{mod}} = \sum_{i=1}^{N_p}\frac{1}{2}\left\|M^i_{\mbox{p}}{\odot}M{\odot}\left(WHRFh_m^i-l_m\right)\right\|_2^2 + \\
\frac{1}{2} \lambda_p \sum_{k{\in}\mathcal{N}_i} \left\|h_m^i-h_m^k\right\|^2
\end{align}
where $i$ corresponds to the index of a patch, ${N_p}$ represents the total number of patches, the mask $M^i_{\mbox{p}}$ takes the value of unity only in the region of the LF image corresponding to the $i$th patch and zero elsewhere, and $\mathcal{N}_i$ denotes the set of indices of neighbors of the $i$th patch. The matrix vector product $Fh_m^i$ can also be replaced by $H_m^if$ when the cost is expressed in terms of $f$.
 
The gradient of the modified cost with respect to sharp scene radiance can be written as
\begin{align}
\nonumber
\label{eqn:fgrad}
\frac{\partial E_{\mbox{mod}}}{\partial f} =\sum_{i=1}^{N_p}{H_m^i}^T{{R}^T}{H^T}{{W}^T}\left(M^i_{\mbox{p}}{\odot} \right. \\
\left. M{\odot}\left(WHRH^i_mf-l_m\right)\right)
\end{align}
The gradient with respect to $i$th motion PSF is given by
\begin{align}
\nonumber
\frac{\partial E_{\mbox{mod}}}{\partial h^i_m} ={F}^T{{R}^T}{H^T}{{W}^T}\left(M^i_{\mbox{p}}{\odot}M{\odot}\left(WHRFh_m-l_m\right)\right) \\ +
 \lambda_p \sum_{k{\in}\mathcal{N}_i} \left(h_m^i-h_m^k\right)
\label{eqn:hgrad}
\end{align}
In our modified algorithm, we use the gradients evaluated according to expressions in equations (\ref{eqn:fgrad}) and (\ref{eqn:hgrad}) while implementing the update steps given in (\ref{eqn:fupdate}) and (\ref{eqn:hupdate}). Furthermore, we use a pyramid-based scheme which facilitates faster convergence.

\begin{figure*}[htb]
\begin{center}
\begin{tabular}{ccccc}
\includegraphics[width=97pt]{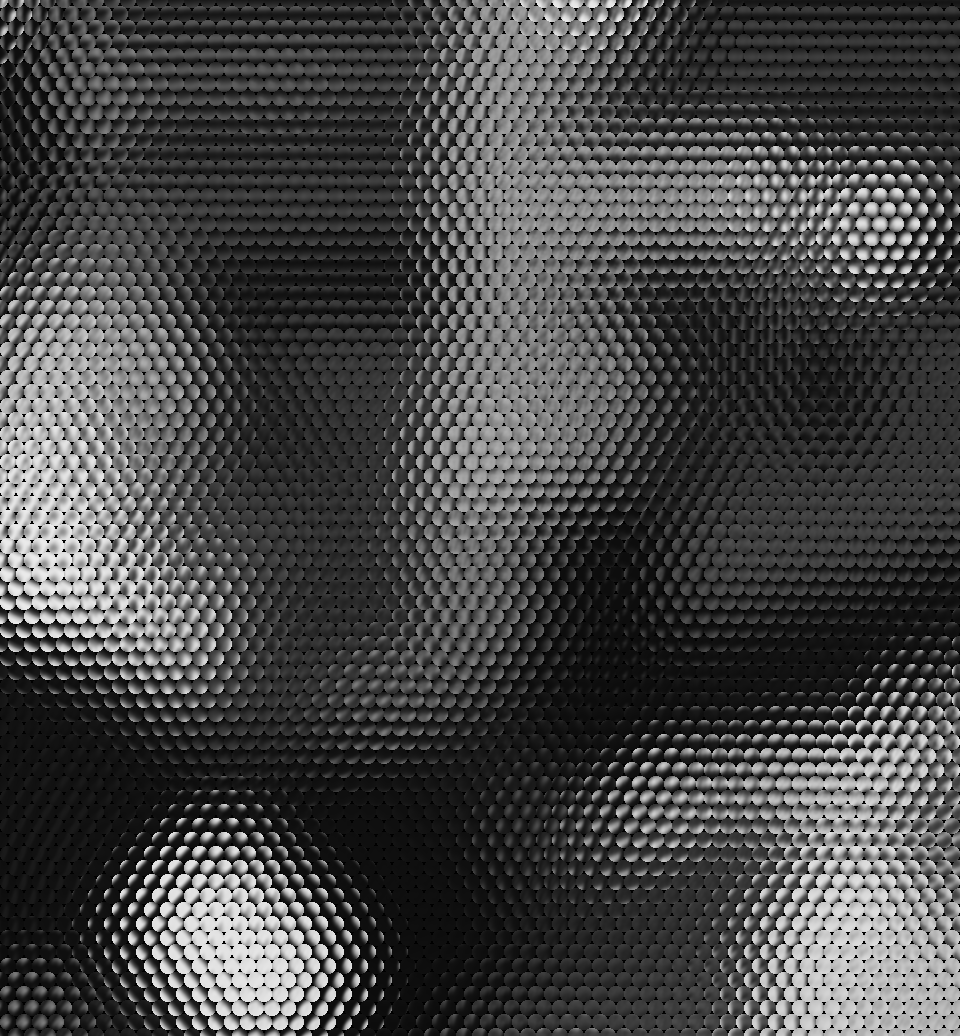}& \hspace{-11pt}
\includegraphics[width=97pt]{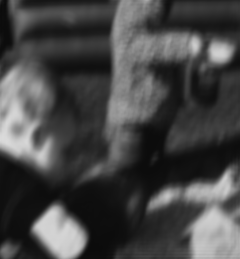}&\hspace{-11pt}
\includegraphics[width=97pt]{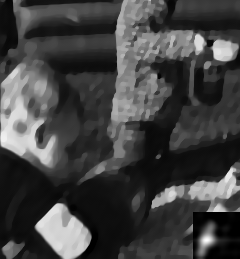}&\hspace{-11pt}
\includegraphics[width=97pt]{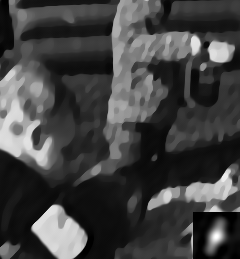}&\hspace{-11pt}
\includegraphics[width=97pt]{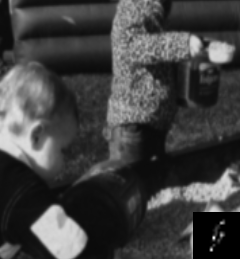} \\
\includegraphics[width=97pt]{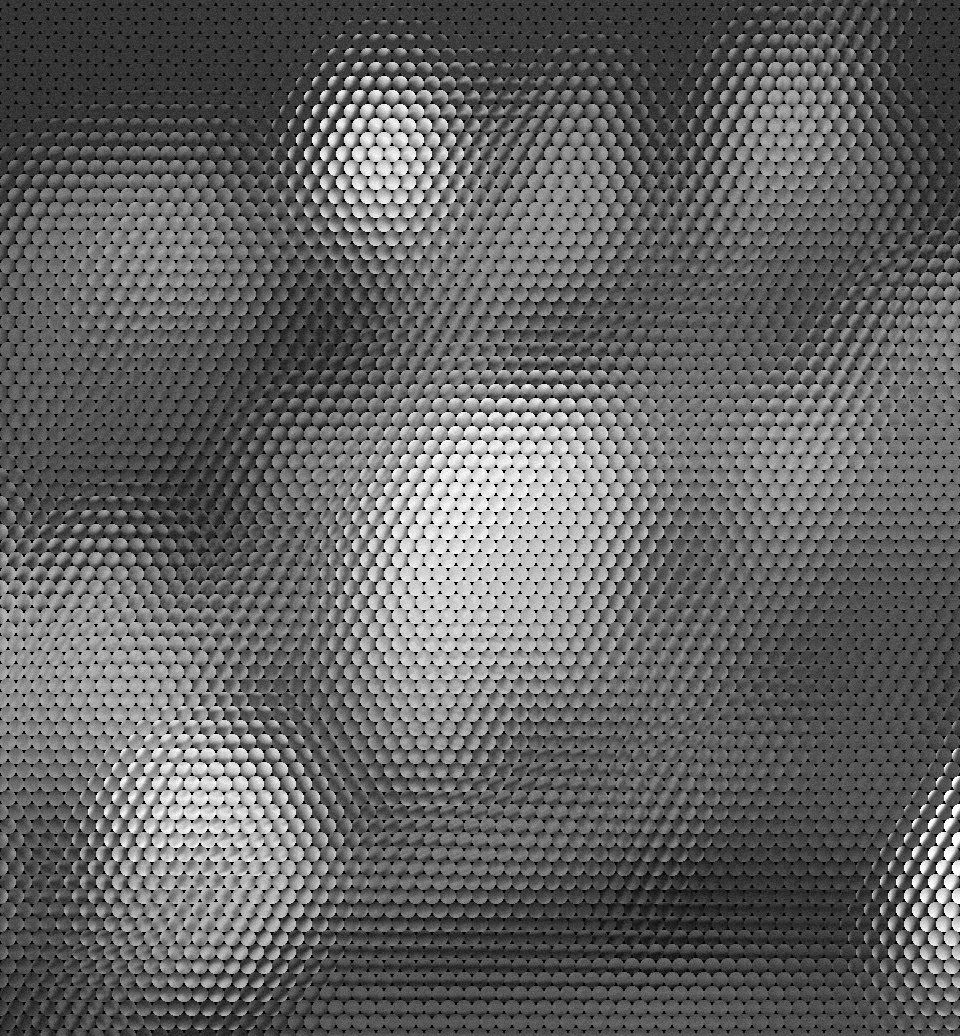}&\hspace{-11pt}
\includegraphics[width=97pt]{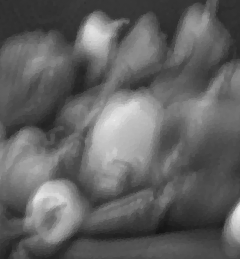}&\hspace{-11pt}
\includegraphics[width=97pt]{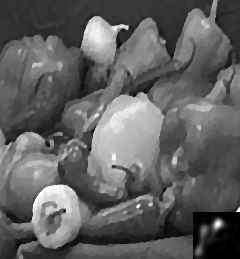}&\hspace{-11pt}
\includegraphics[width=97pt]{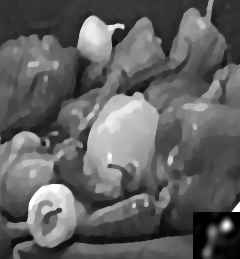}&\hspace{-11pt}
\includegraphics[width=97pt]{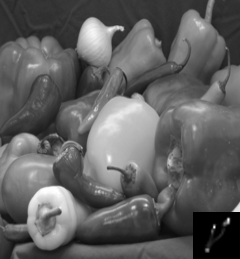} \\
\includegraphics[width=97pt]{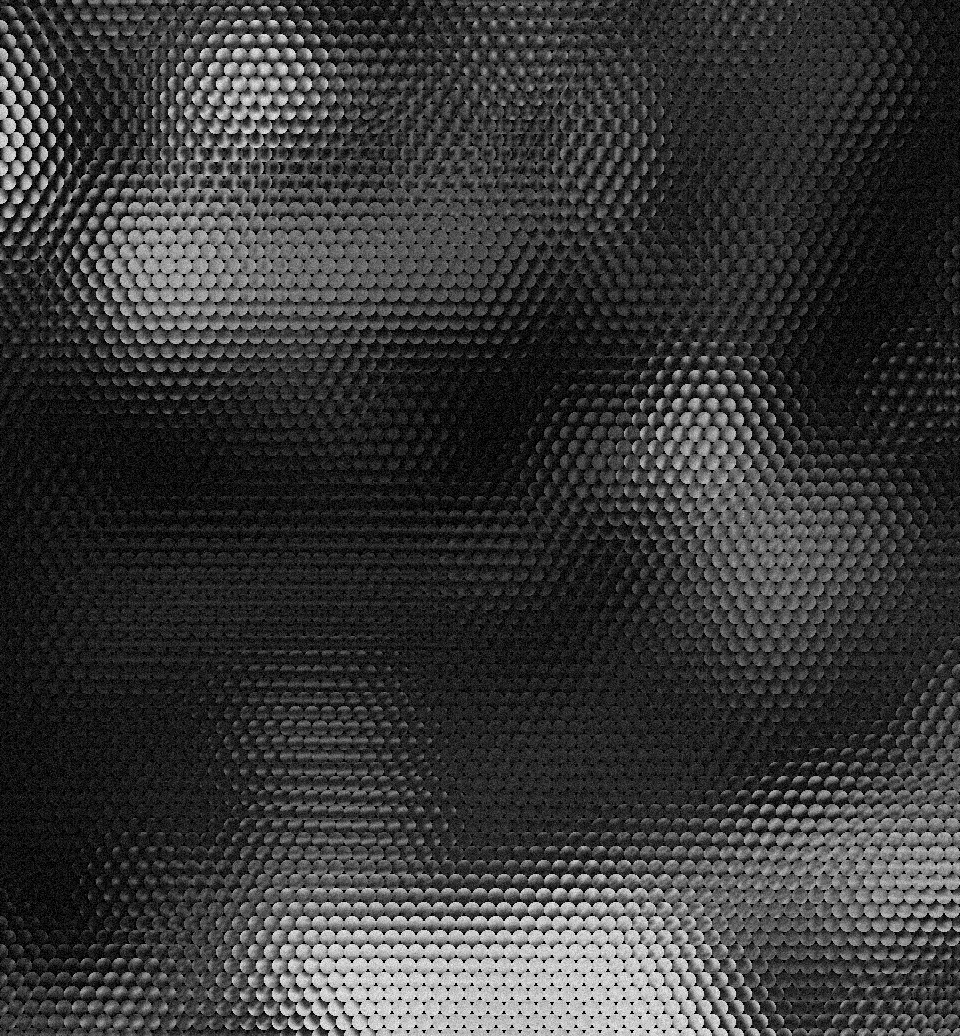}&\hspace{-11pt}
\includegraphics[width=97pt]{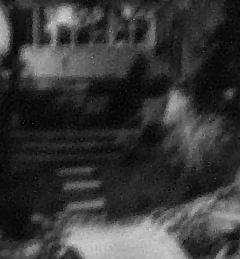}&\hspace{-11pt}
\includegraphics[width=97pt]{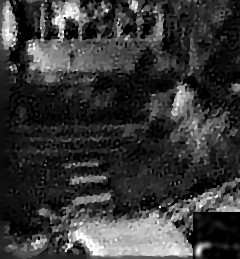}&\hspace{-11pt}
\includegraphics[width=97pt]{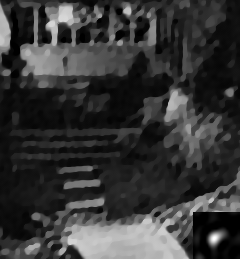}&\hspace{-11pt}
\includegraphics[width=97pt]{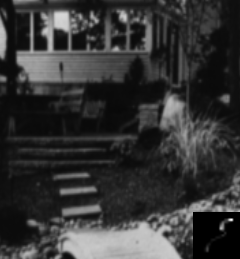} \\
\end{tabular}
\caption{Example images from the synthetic experiment. First column: LF images, second column: texture without deconvolution, third column: \texttt{two-step} results, fourth column: results of the proposed algorithm, and fifth column: true PSF and texture. 
\label{fig:synth}}
\end{center}
\end{figure*}

\begin{table*}[htb]
\centering
\begin{tabular}{|c*{13}{|c}| }
 \hline 
 \multicolumn{5}{@{}|c|@{}}{\bf Noise-free} &\multicolumn{4}{@{}|c|@{}}{\bf $2.5 \%$  noise}&\multicolumn{4}{@{}|c|@{}}{\bf $5 \%$  noise} \\ 
  \hline
 &  \multicolumn{2}{@{}c|@{}}{PSNR} &   \multicolumn{2}{@{}c|@{}}{SSIM} &  \multicolumn{2}{@{}c|@{}}{PSNR} &   \multicolumn{2}{@{}c|@{}}{SSIM}&  \multicolumn{2}{@{}c|@{}}{PSNR} &   \multicolumn{2}{@{}c|@{}}{SSIM}  \\ 
 &      \multicolumn{1}{@{}c@{}} {\tiny \bf ~~proposed~~}  &   \multicolumn{1}{@{}c|@{}} {\tiny \bf ~~two-step~~}   &  \multicolumn{1}{@{}c@{}} {\tiny \bf ~~proposed~~}  &   \multicolumn{1}{@{}c|@{}} {\tiny \bf ~~two-step~~}   &       \multicolumn{1}{@{}c@{}} {\tiny \bf ~~proposed~~}  &   \multicolumn{1}{@{}c|@{}} {\tiny \bf ~~two-step~~}    &  \multicolumn{1}{@{}c@{}} {\tiny \bf ~~proposed~~}  &   \multicolumn{1}{@{}c|@{}} {\tiny \bf ~~two-step~~} &       \multicolumn{1}{@{}c@{}} {\tiny \bf ~~proposed~~}  &   \multicolumn{1}{@{}c|@{}} {\tiny \bf ~~two-step~~}    &  \multicolumn{1}{@{}c@{}} {\tiny \bf ~~proposed~~}  &   \multicolumn{1}{@{}c|@{}} {\tiny \bf ~~two-step~~} \\ 
\hline 
  ~ {\tiny$\mu$} ~ &\tiny{ {\bf 28.51}} &  \tiny{ 27.13 } &\tiny{{\bf 0.873}}  & \tiny{ 0.821}  &\tiny{  {\bf28.27}}  & \tiny{ 26.43}  & \tiny{ {\bf0.864} }& \tiny{  0.797}&\tiny{  {\bf25.31}}  & \tiny{ 21.98}  & \tiny{ {\bf0.676} }& \tiny{  0.521} \\
  \tiny{$\sigma$}   & \tiny{0.83} &  \tiny{4.71}   &  \tiny{0.021} &  \tiny{0.173}   &  \tiny{0.850} &    \tiny{4.02}  &   \tiny{0.021} &  \tiny{ 0.154}  &  \tiny{0.961} &    \tiny{2.93}  &   \tiny{0.035} &  \tiny{ 0.105}\\ 
  \hline 
 \end{tabular}
 \vspace{0.5cm}
\caption{In this table we show a comparison between the methods. We show the average ($\mu$) and the standard deviation ($\sigma$) of the PSNR and SSIM metrics for all synthetically generated motion blurred light field images. } \label{tab:tabledeblurring}
\vspace{-0.2cm}
\end{table*}

\section{Experimental results}
We evaluate our light field blind deconvolution algorithm on several synthetic and real experiments. For synthetic experiments, we assume ideal imaging conditions as in section \ref{sec:lfbd}. We artificially generate motion blurred light field image by assuming scene texture, depth, camera parameters, and motion blur kernel. We perform real experiments with motion blurred observation captured from a handheld Lytro Illum camera. In both synthetic as well as real experiments, we apply our alternate minimization scheme to solve for the sharp texture and motion PSF with higher TV regularization. Subsequently, with the estimated PSF, we solve for the sharp texture alone with lesser TV regularization. For the purpose of comparison, we also show results obtained by a \texttt{two-step} approach. i.e., first the texture is estimated from the LF image without compensating for motion blur. The sharp image is then estimated by following conventional blind deconvolution techniques.

\subsection{Synthetic experiments}
We performed synthetic experiments on a database consisting of $12$ pairs of PSFs and scene textures. For scene textures, we used \texttt{peppers.png} image of Matlab and two images from the dataset of Levin et al.~\cite{Levin2011Understanding}. For the motion blur kernels, we chose four different PSFs from \cite{Levin2011Understanding}. We assumed realistic camera parameters and depth values to calculate the LF PSF.  The values of all the parameters involved are summarized in Table~\ref{tab:settings}. For each combination of motion blur and texture, we generated the LF image and added $0\%$, $2.5\%$ and $5\%$ zero mean white Gaussian noise. We compared the results of our method to the output of \texttt{two-step} approach wherein the algorithm in \cite{Perrone2014} was used for blind deconvolution. The value of TV regularization parameter used was the same for both the approaches. For the noise-free experiments we set the regularization weight as $\lambda = 1.6\cdot 10^{-3}$ for the alternating minimization scheme, and as $\lambda = 7\cdot 10^{-4}$ to perform the final deblurring step. For the experiments with noisy images we use the values $\lambda = 2\cdot 10^{-3}$, and $\lambda = 8\cdot 10^{-4}$ for regularization in the initial and final stages, respectively. Note that in all our experiments, the input light field images as well as the textures take values between $0$ and $1$. In our comparison we used the Peak Signal-to-Noise Ratio (PSNR) and the Structural Similarity Index (SSIM)~\cite{Wang2004} metrics. Since between the blur function and the sharp image there is a translational ambiguity, for each image we take the maximum PSNR and SSIM among all the possible shifts between the estimate and ground truth image. In Table~\ref{tab:tabledeblurring} we show the mean and standard deviation of PSNR and SSIM for the results of the proposed approach and the \texttt{two-step} method. For noise-free as well as noisy scenarios, the proposed method outperforms the \texttt{two-step} method for both metrics. The \texttt{two-step} method also has a larger standard deviation. In contrast, the proposed algorithm has a small standard deviation and therefore is more consistent and robust to noise.

We show representative examples of our experiment in Fig. \ref{fig:synth}. While the images in the first row corresponds to the no-noise scenario, those in the second and third row correspond to the scenarios of $2.5\%$ and $5\%$ noise, respectively. The true texture and motion PSF (inserted at the bottom right corner) shown in the last column were used to generate the LF image shown in the first column. The results of the \texttt{two-step} and proposed methods are shown in third and fourth columns, respectively. For visual comparison, in the second column, we show the result of estimating the texture without compensating for motion blur (i.e., the output of the first step of the \texttt{two-step} approach). Images in Fig. \ref{fig:synth} indicate that the performance of the proposed method is quite good.

\begin{table}[b]
\centering
\begin{tabular}{|l|r| }
\hline
pixels per microlens & 16 $\times$ 16\\
vertical spacing between even rows & 28 pixels\\
main lens focal length ($F$) & 0.0095\\
pixel size & 1.4 $\mu$m\\
main lens F-number & 2.049\\
microlens spacing & 20 $\mu$m\\
main lens to microlens array distance & 9.8 mm\\
microlens array to sensor distance & 47.8 $\mu$m\\
microlens focal length & 48.0 $\mu$m\\
\hline
\end{tabular}
\vspace{.5cm}
\caption{Camera parameters.\label{tab:settings}}
\end{table}

 \begin{figure*}[htb!]
\begin{center}
\begin{tabular}{ccc}
\includegraphics[width=167pt]{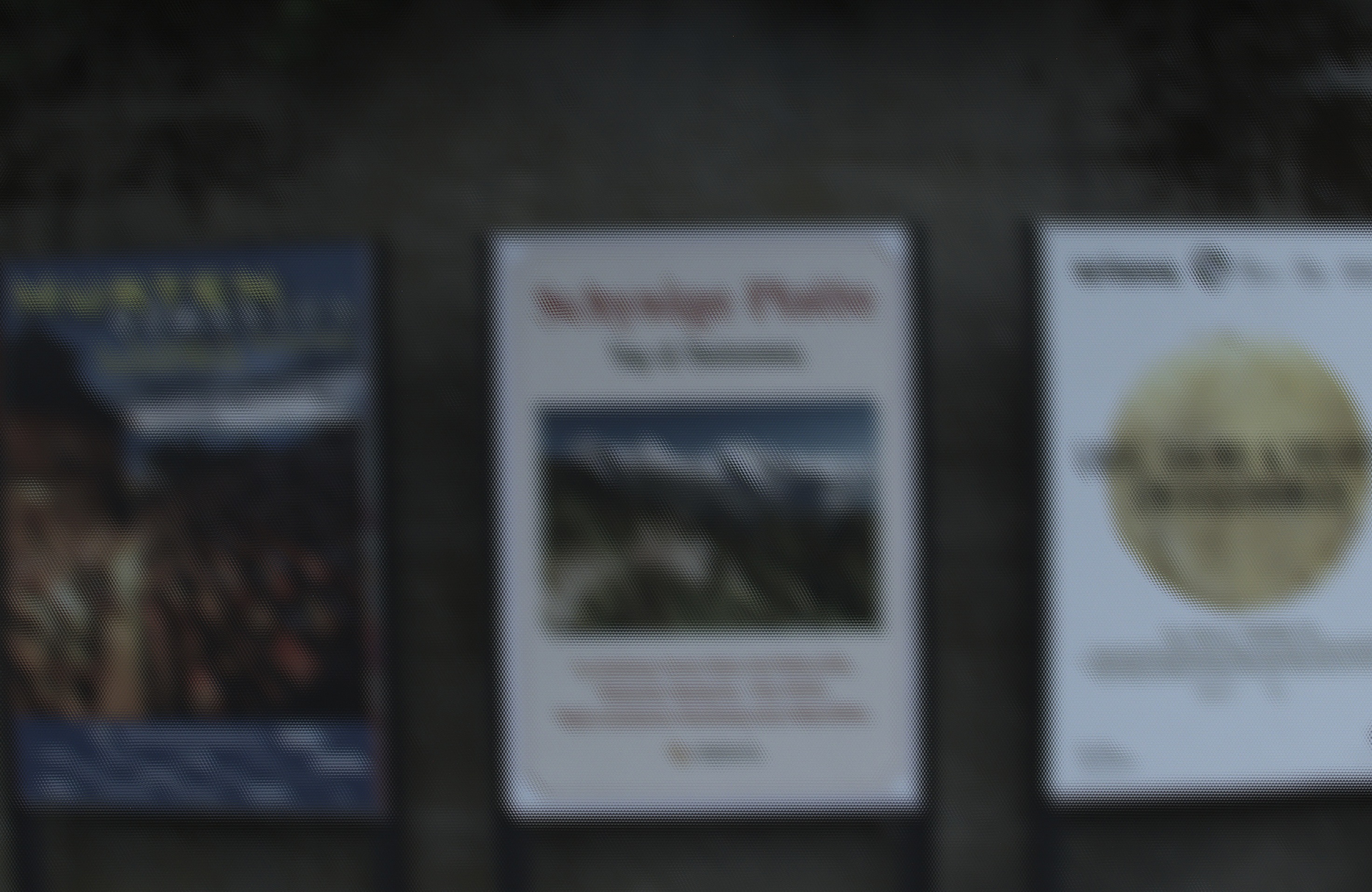}&\hspace{-12pt}
\includegraphics[width=167pt]{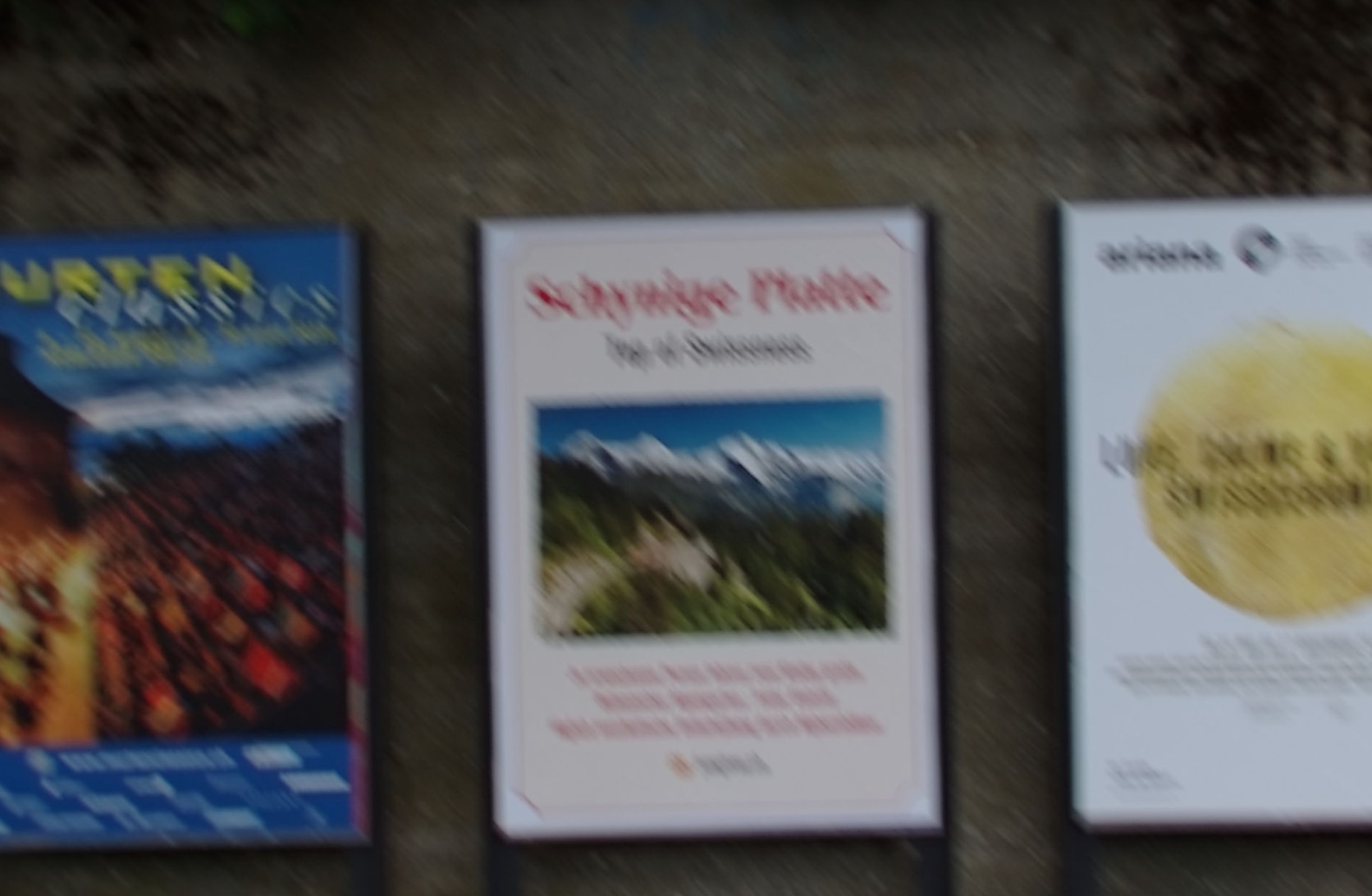}&\hspace{-12pt}
\includegraphics[width=167pt]{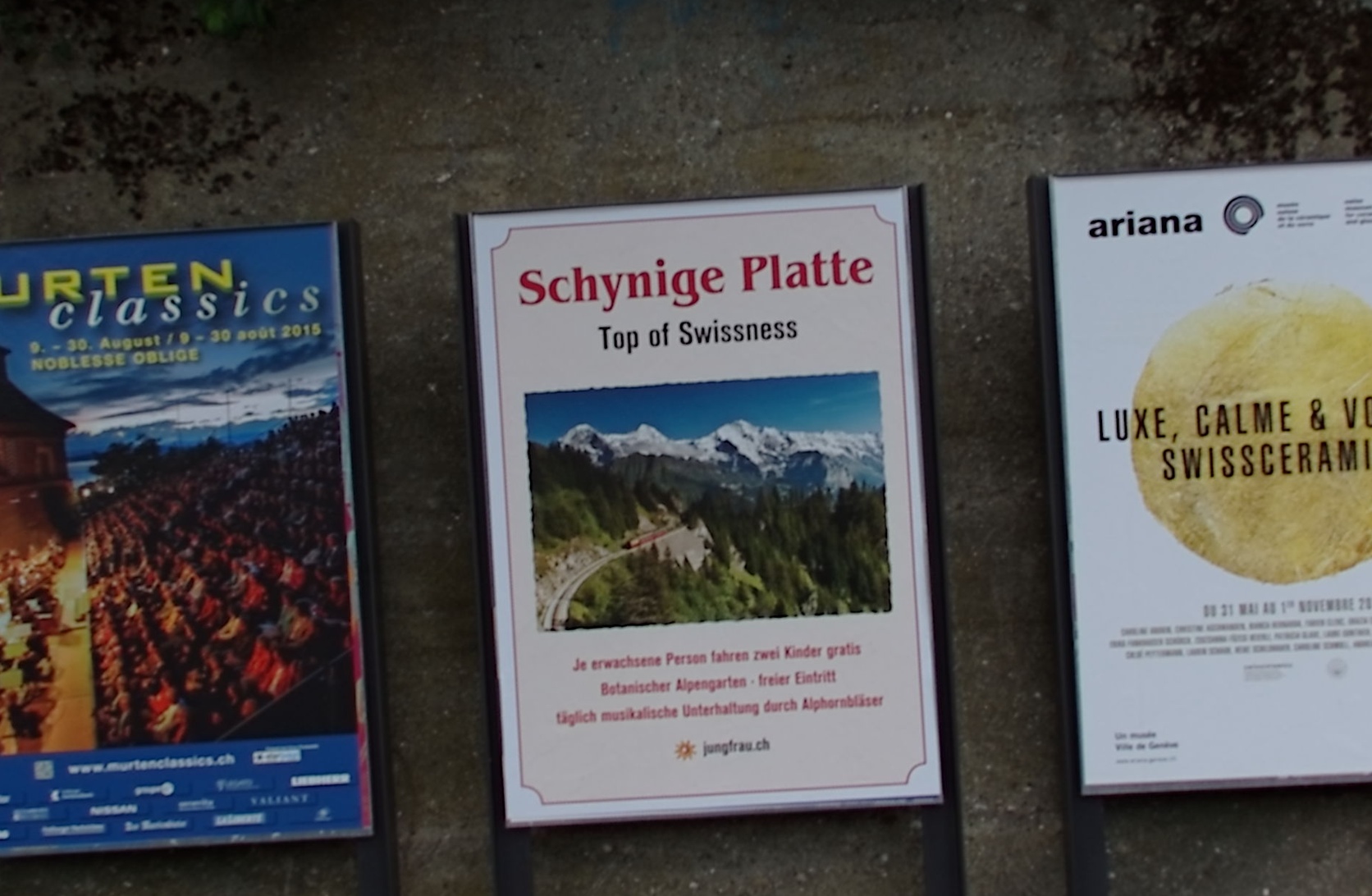}\\
\footnotesize{Raw LF image}&\footnotesize{Refocused image rendered by Lytro software}&\footnotesize{Reference observation without motion blur}\\
\includegraphics[width=167pt]{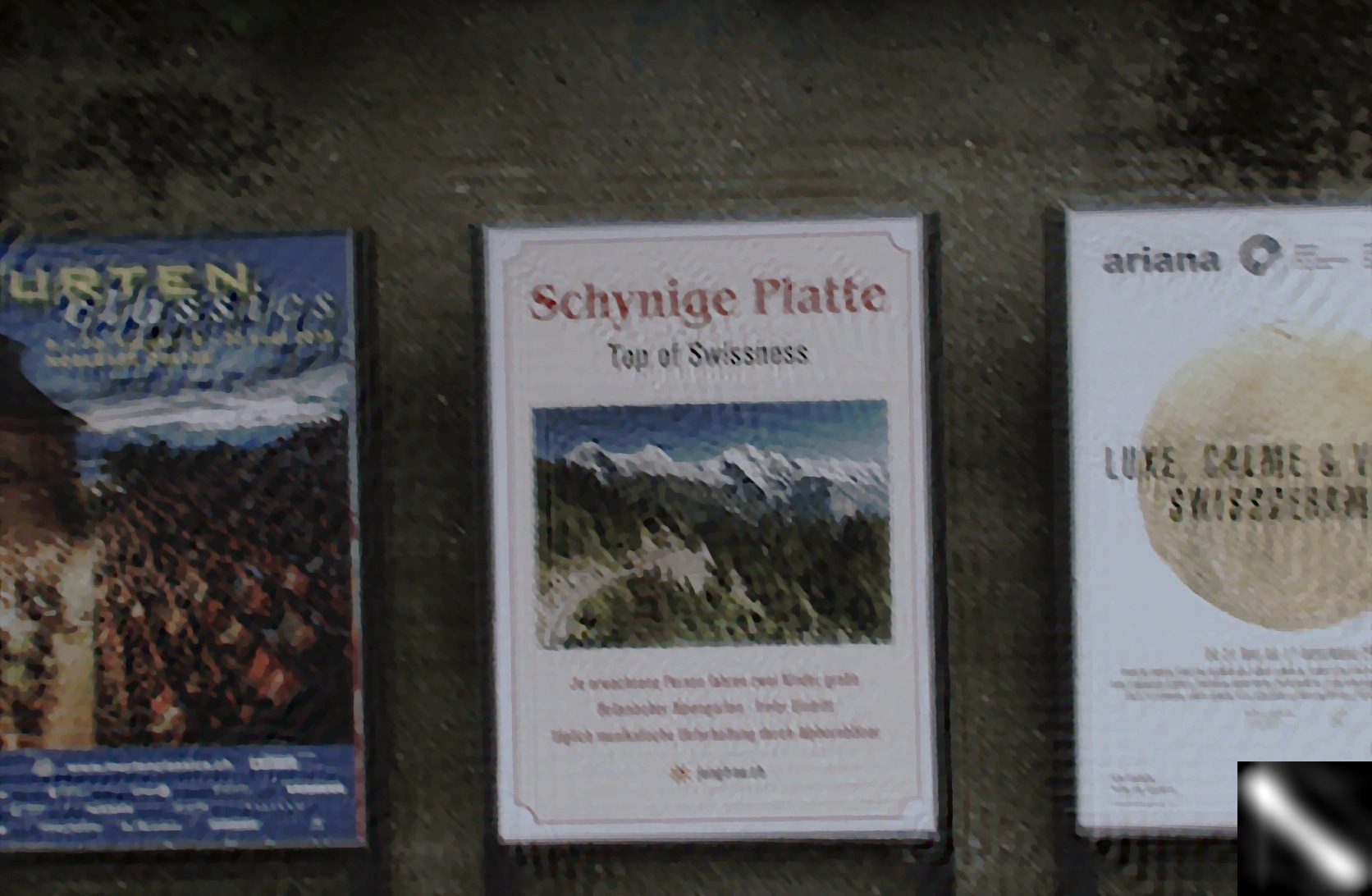}& \hspace{-12pt}
\includegraphics[width=167pt]{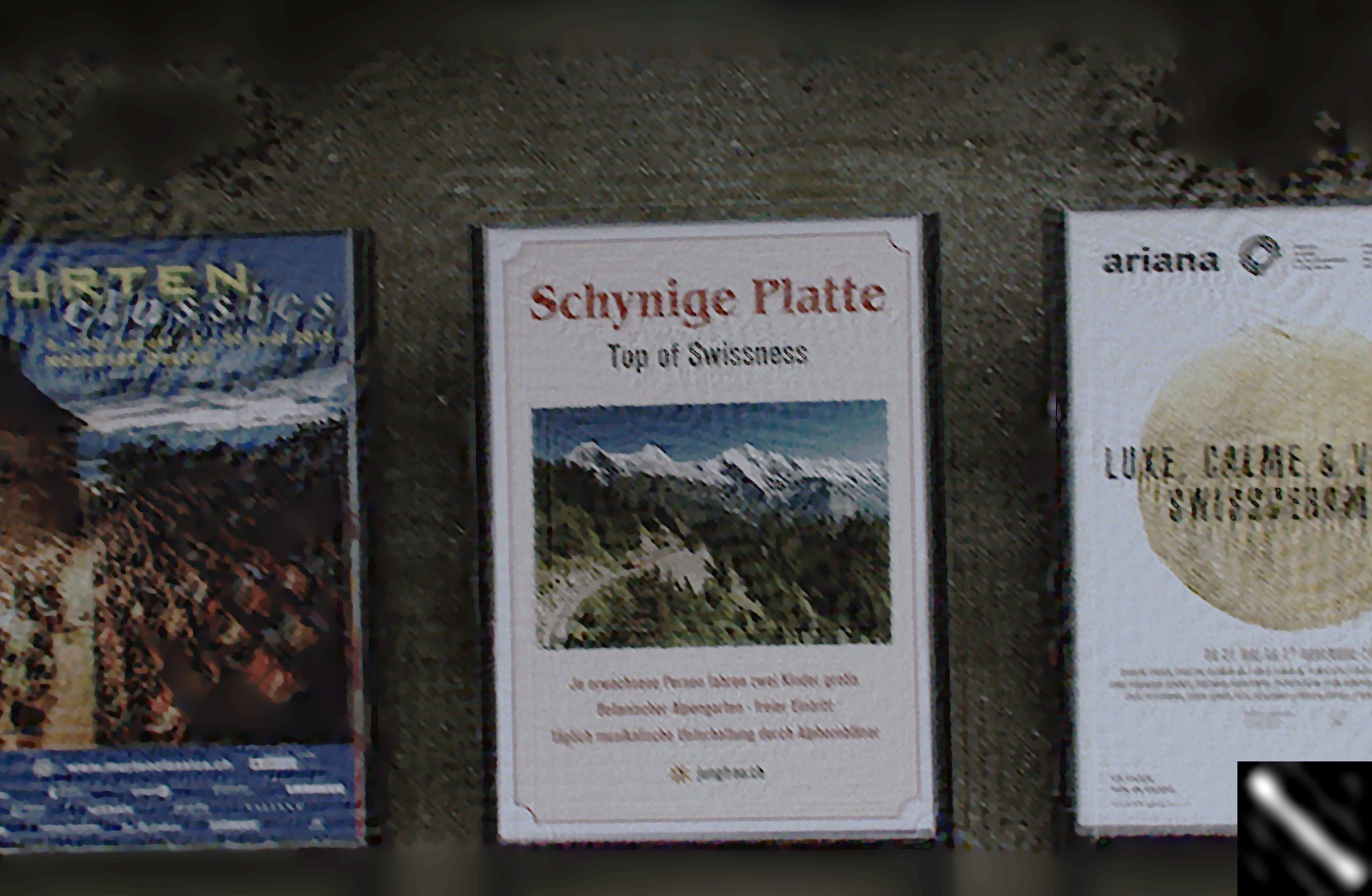}&\hspace{-12pt}
\includegraphics[width=167pt]{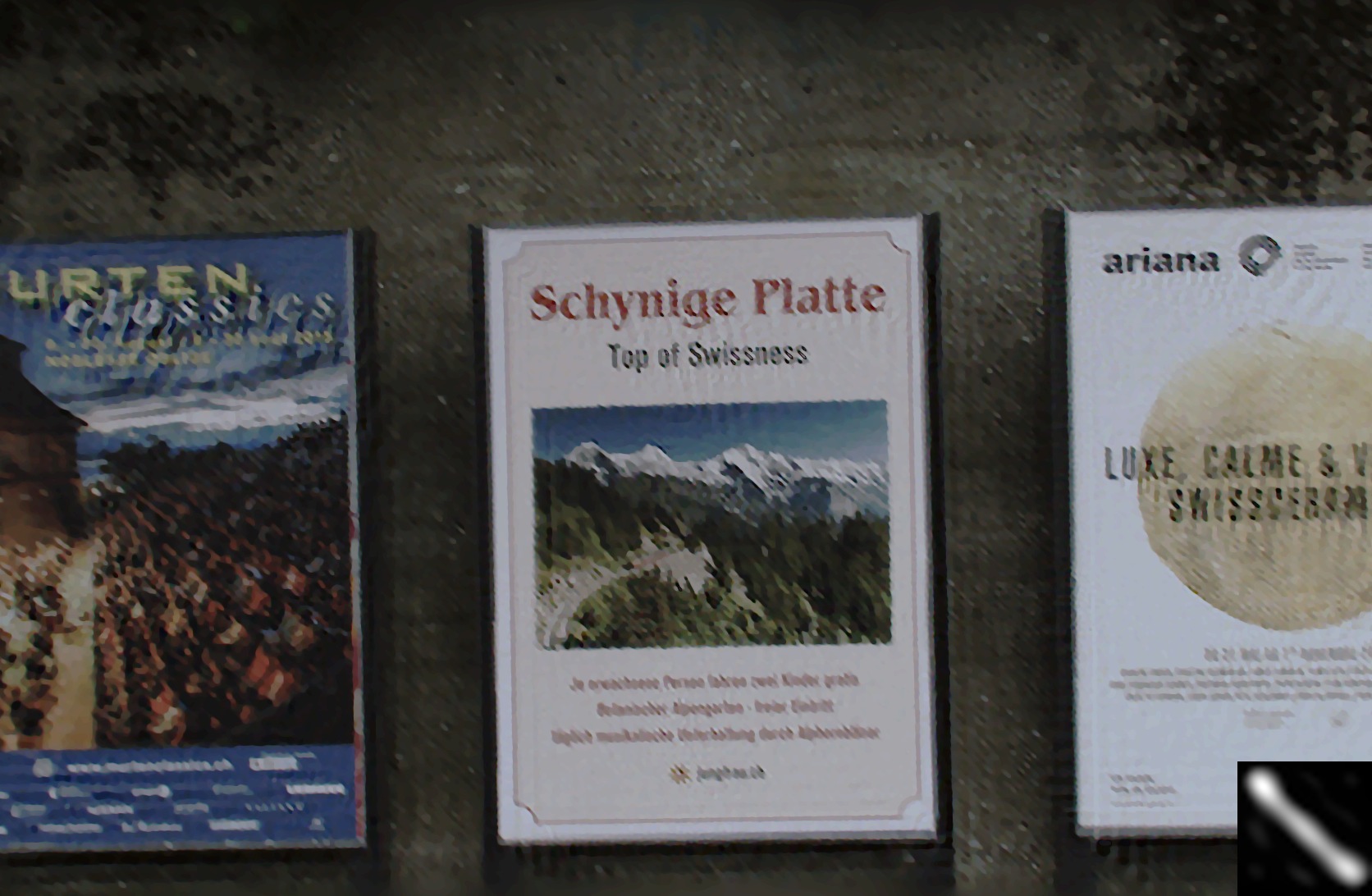}\\
\footnotesize{\texttt{two-step} output}&\footnotesize{Uniform motion blur approach}&\footnotesize{Proposed unified framework}\\
\includegraphics[width=167pt]{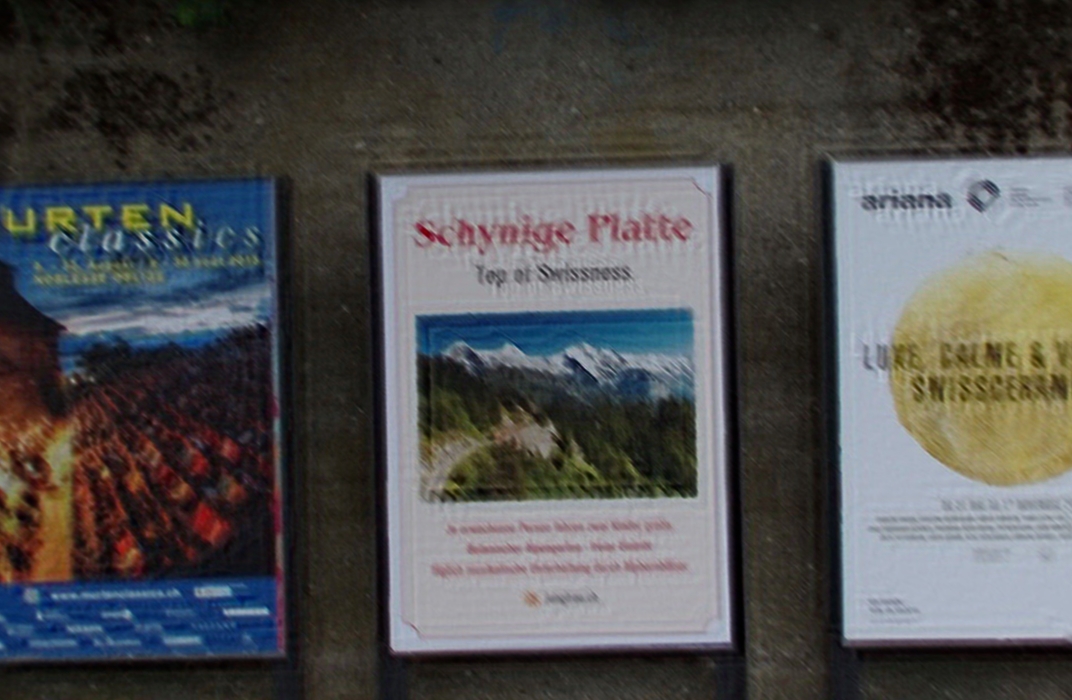}&\hspace{-12pt}
\includegraphics[width=167pt]{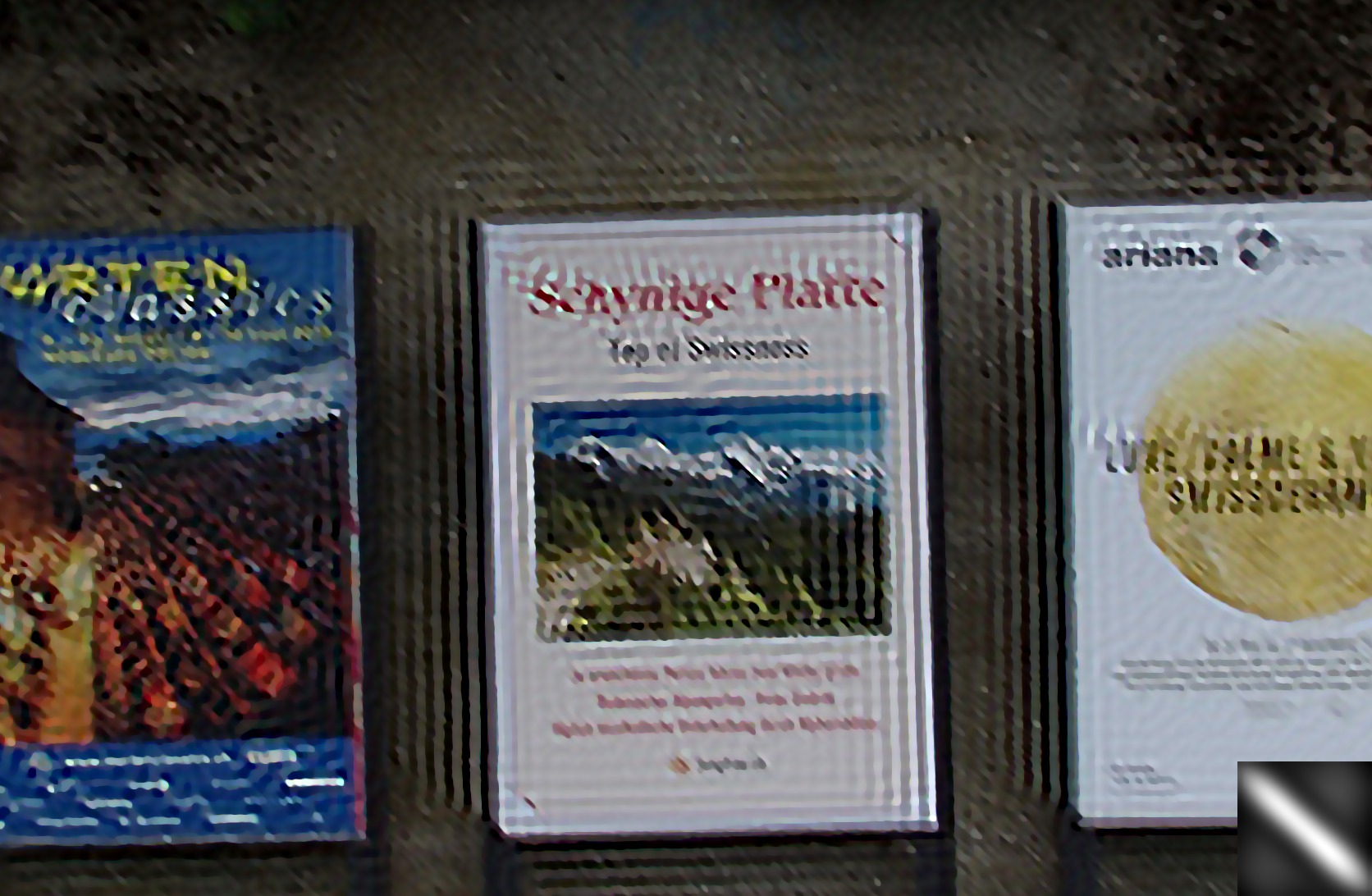}&\hspace{-12pt}
\includegraphics[width=167pt]{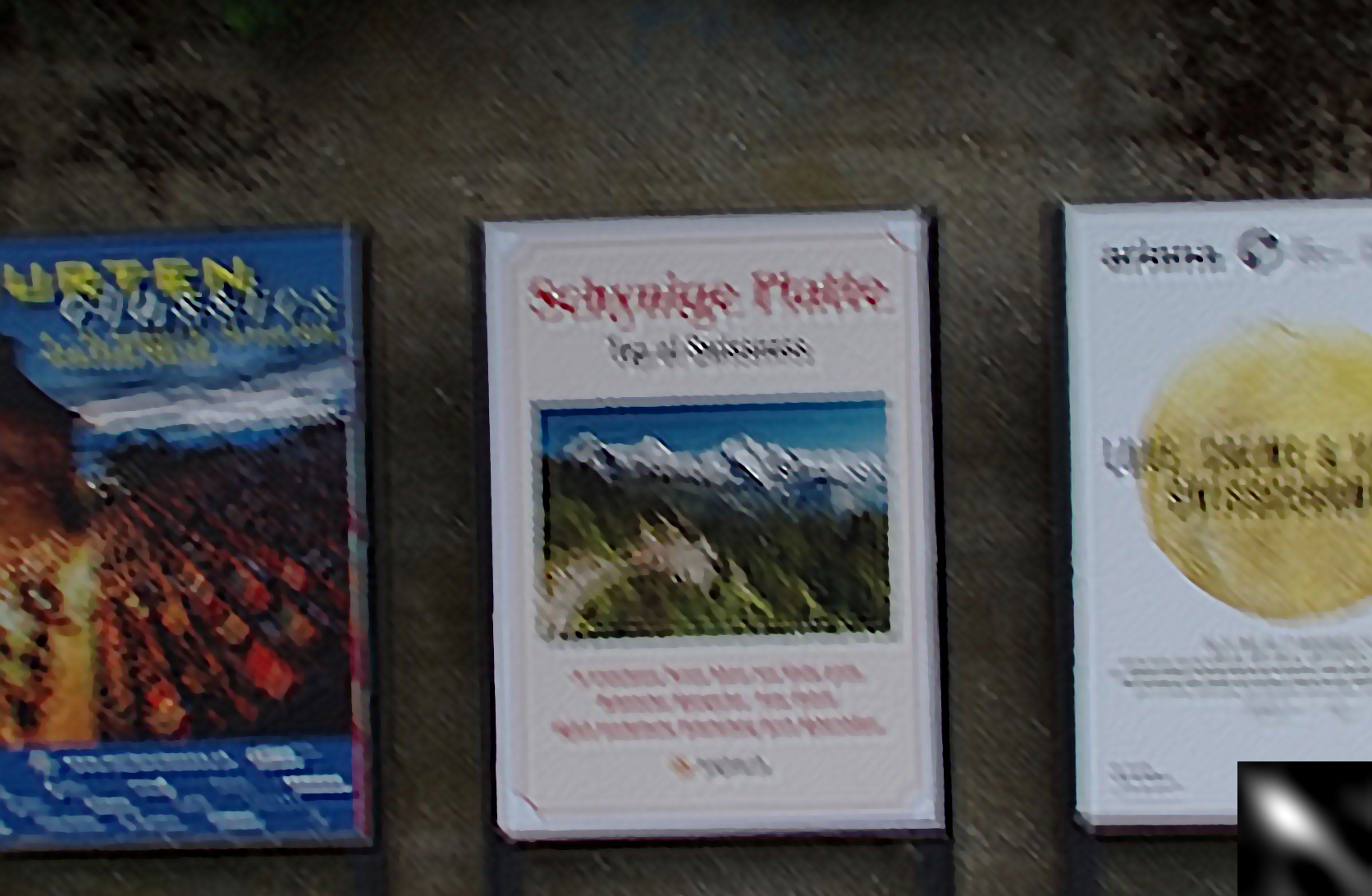}\\
\footnotesize{Output of \cite{ppr:whyte}} on Lytro rendered image &\footnotesize{Using PSF estimated from \cite{babacan}}&\footnotesize{Using PSF estimated from \cite{logtv}}\\
\end{tabular}
\end{center}
\caption{Deblurring results on image of a poster with PSF shown in the insets
\label{fig:post}}
\end{figure*}

\begin{figure*}[htb!]
\begin{center}
\begin{tabular}{ccc}
\includegraphics[width=167pt]{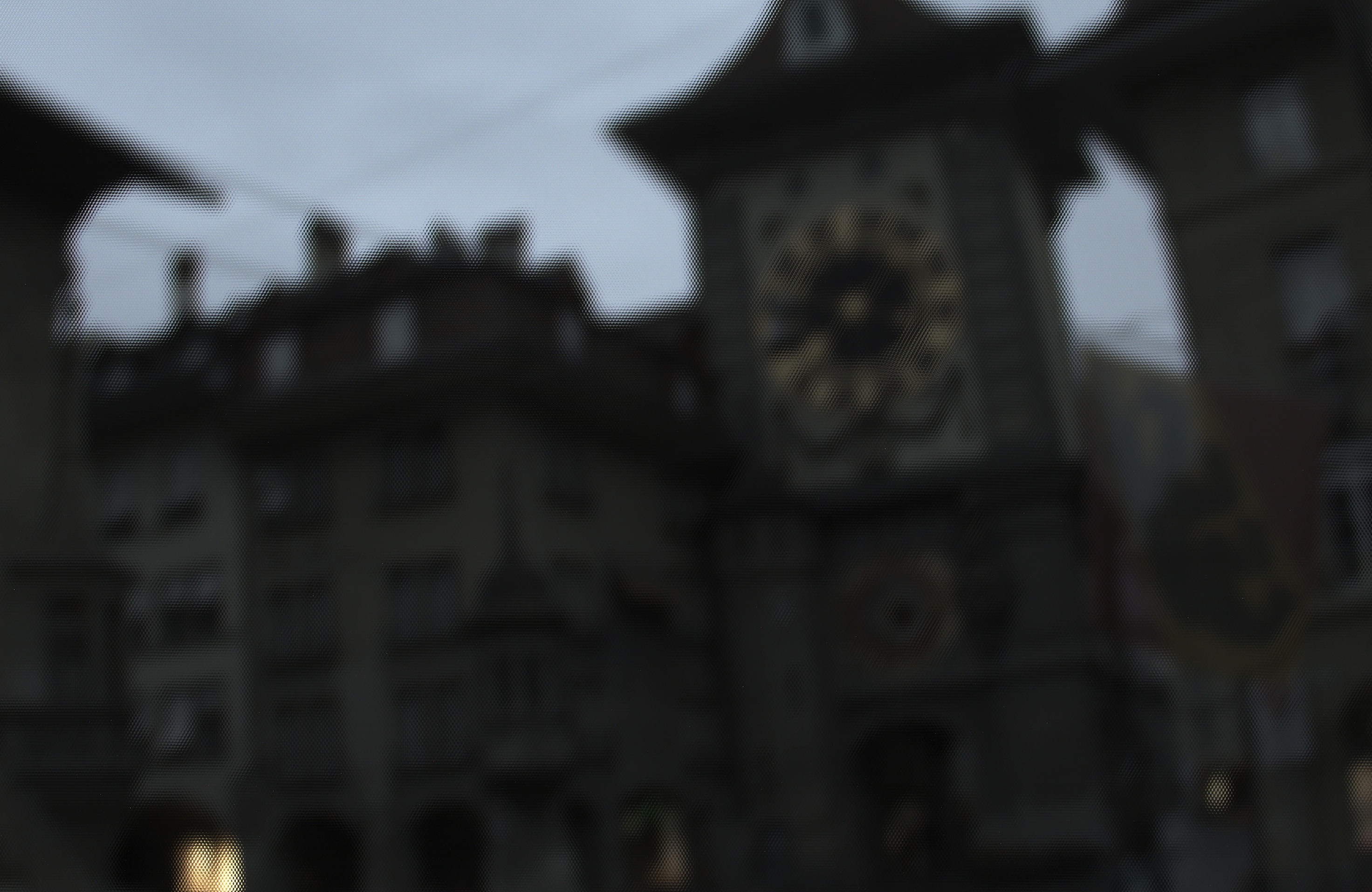}&\hspace{-12pt}
\includegraphics[width=167pt]{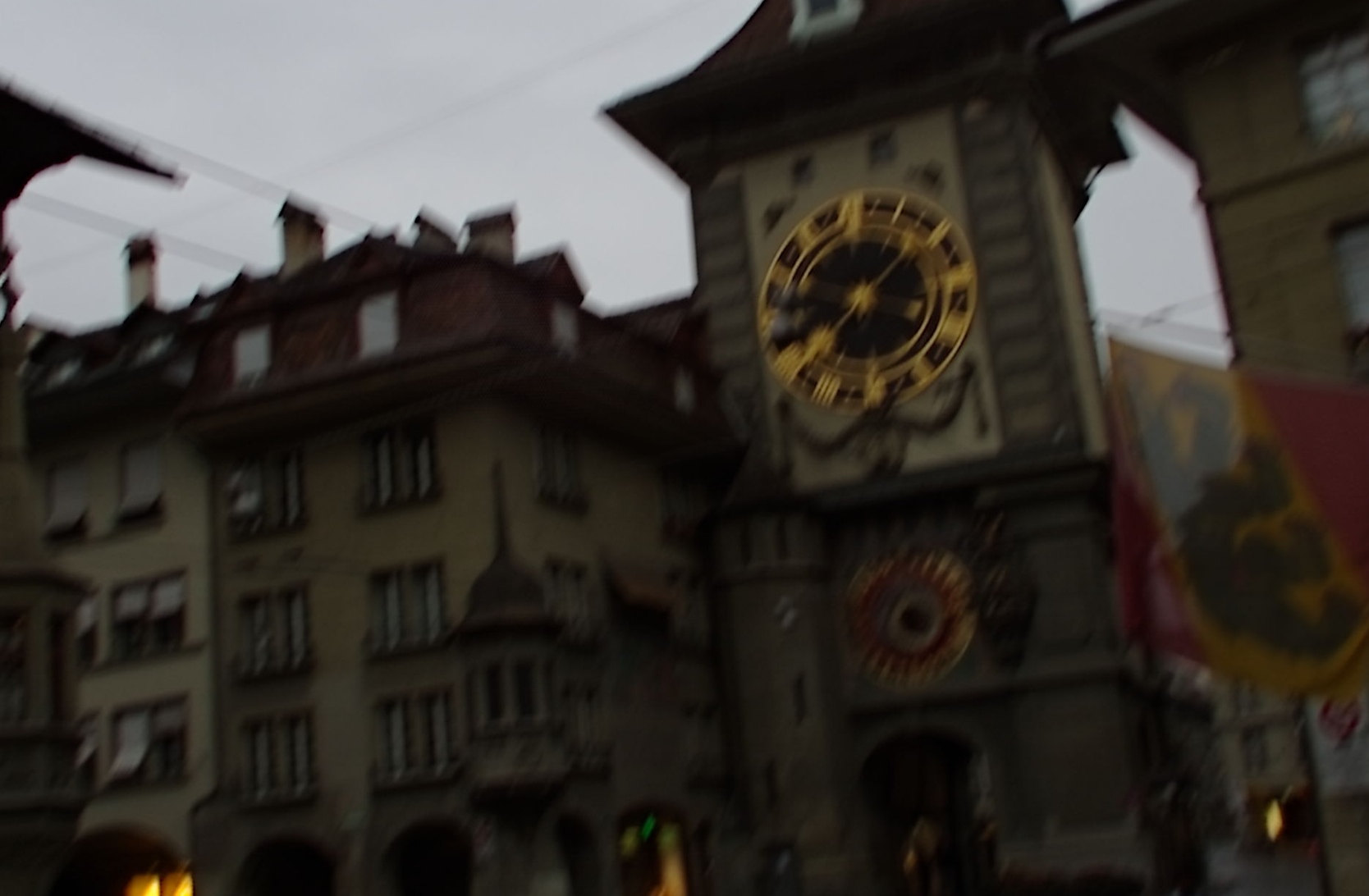}&\hspace{-12pt}
\includegraphics[width=167pt]{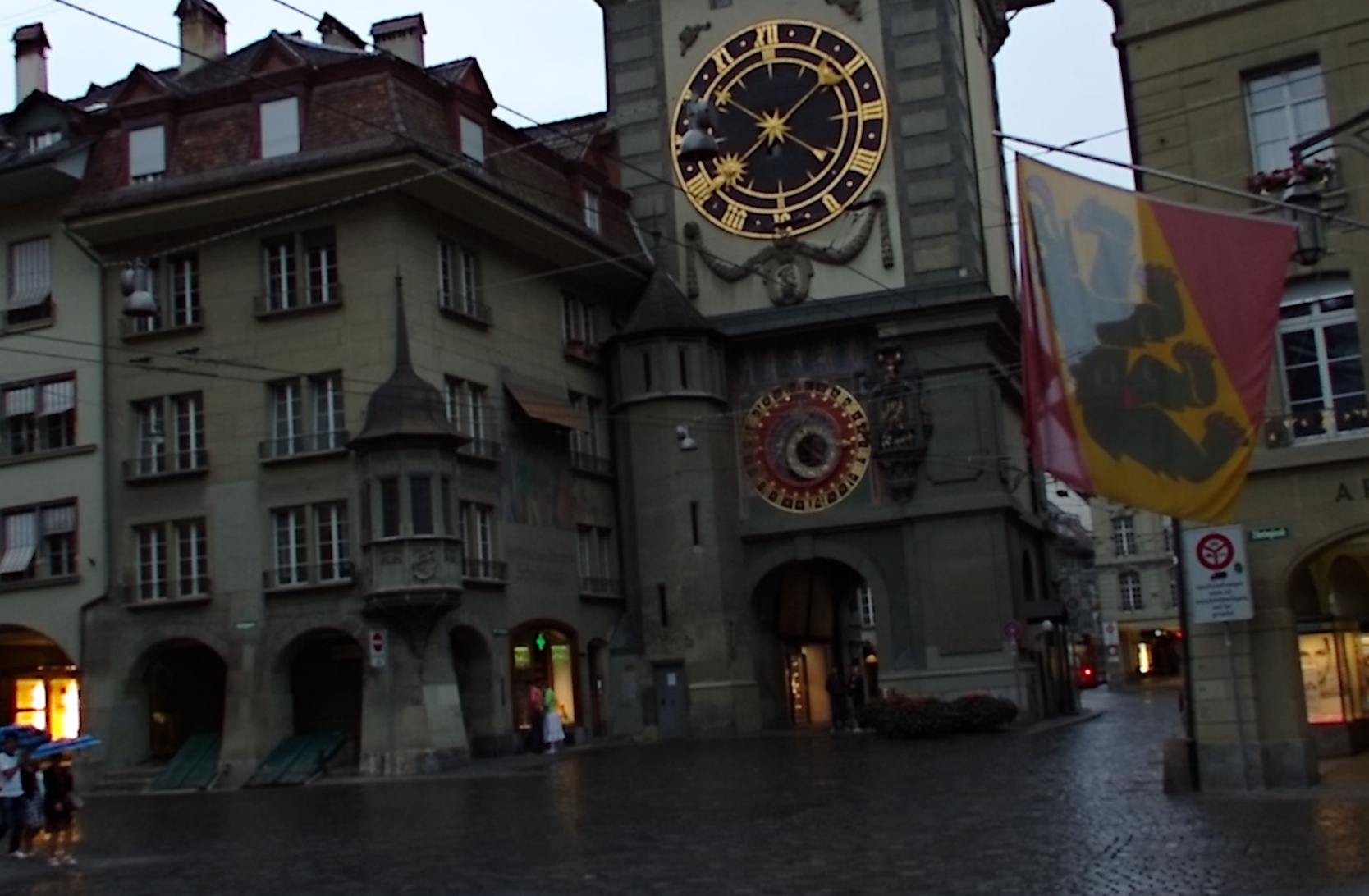}\\
\footnotesize{Raw LF image}&\footnotesize{Refocused image rendered by Lytro}&\footnotesize{Reference observation without motion blur}\\
\includegraphics[width=167pt]{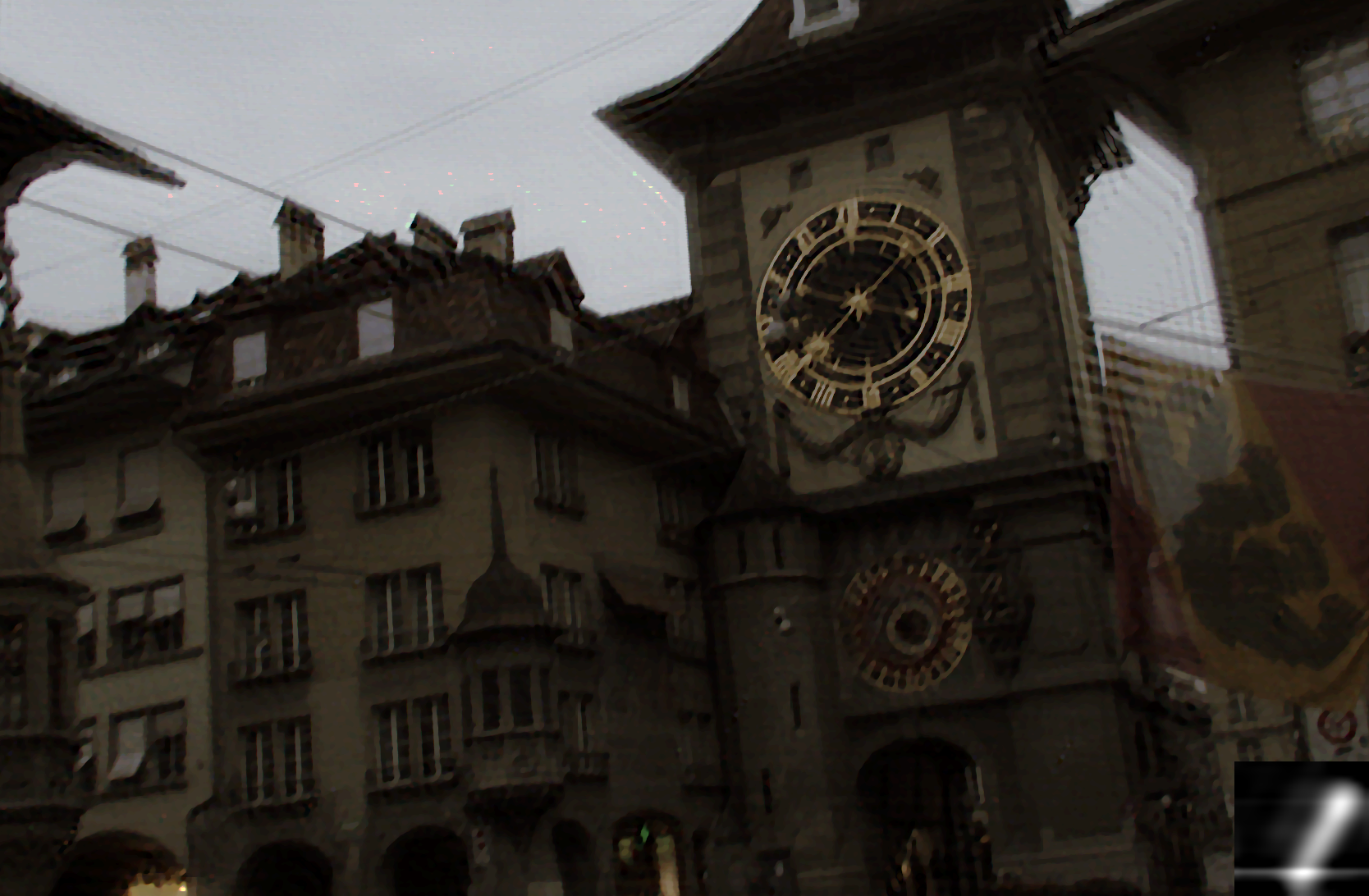}& \hspace{-12pt}
\includegraphics[width=167pt]{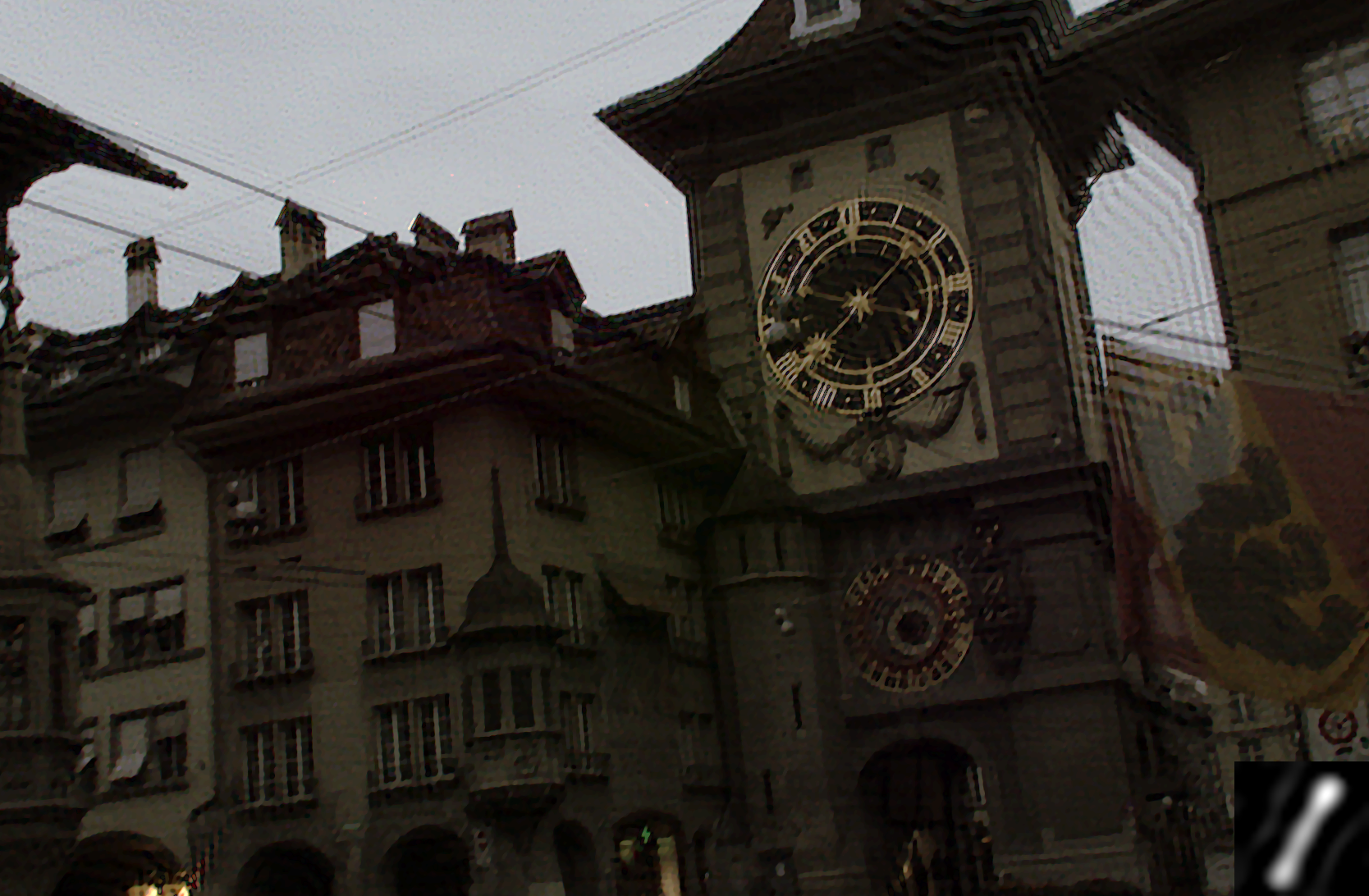}&\hspace{-12pt}
\includegraphics[width=167pt]{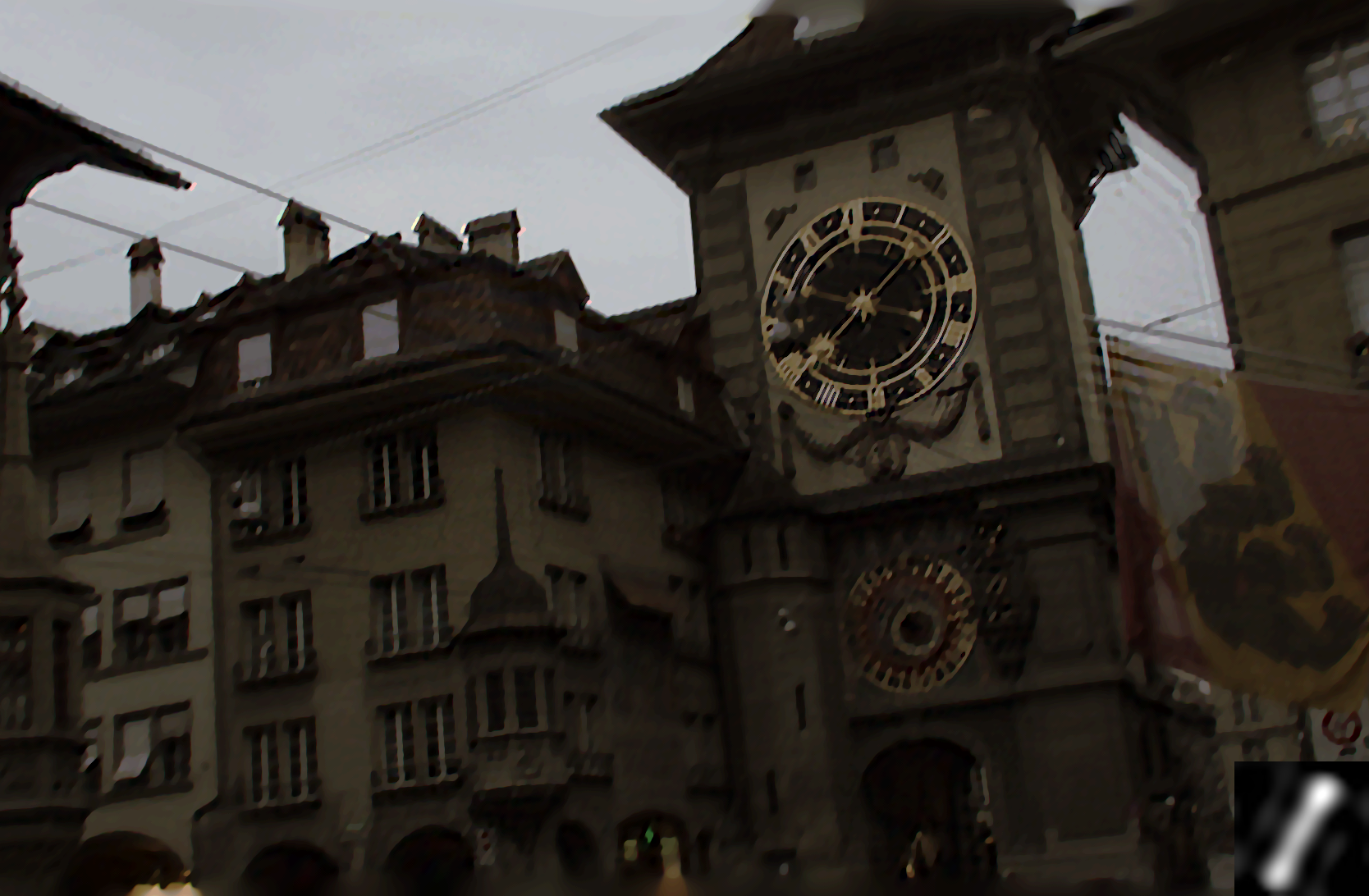}\\
\footnotesize{\texttt{two-step} output}&\footnotesize{Uniform motion blur approach}&\footnotesize{Proposed unified framework}\\
\includegraphics[width=167pt]{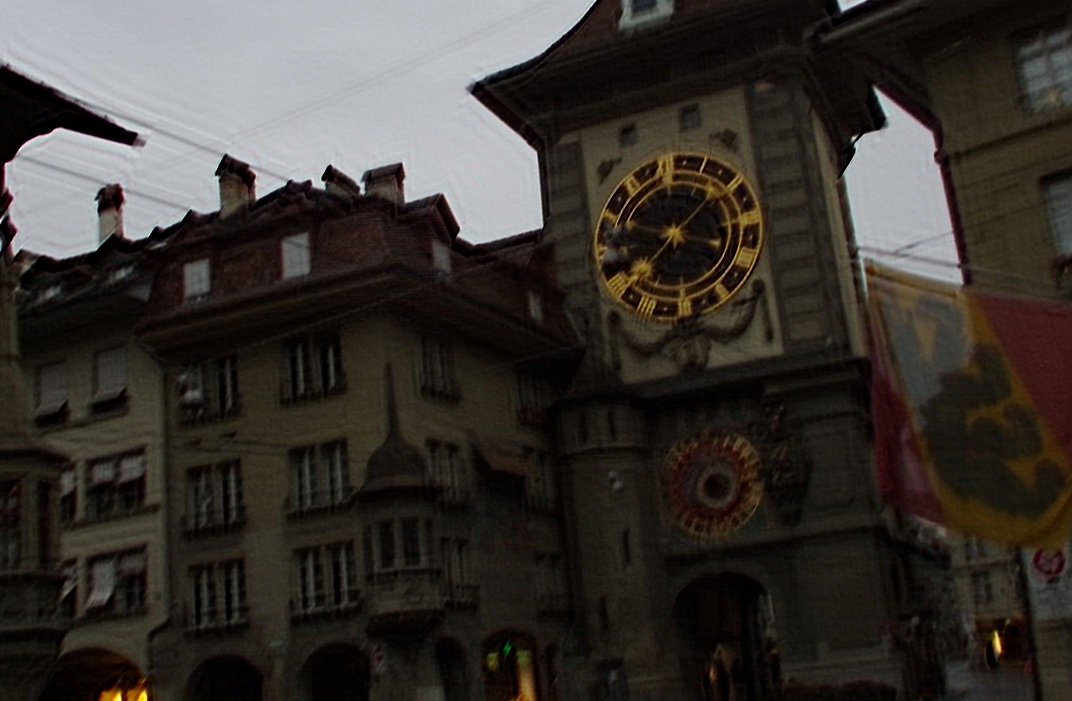}&\hspace{-12pt}
\includegraphics[width=167pt]{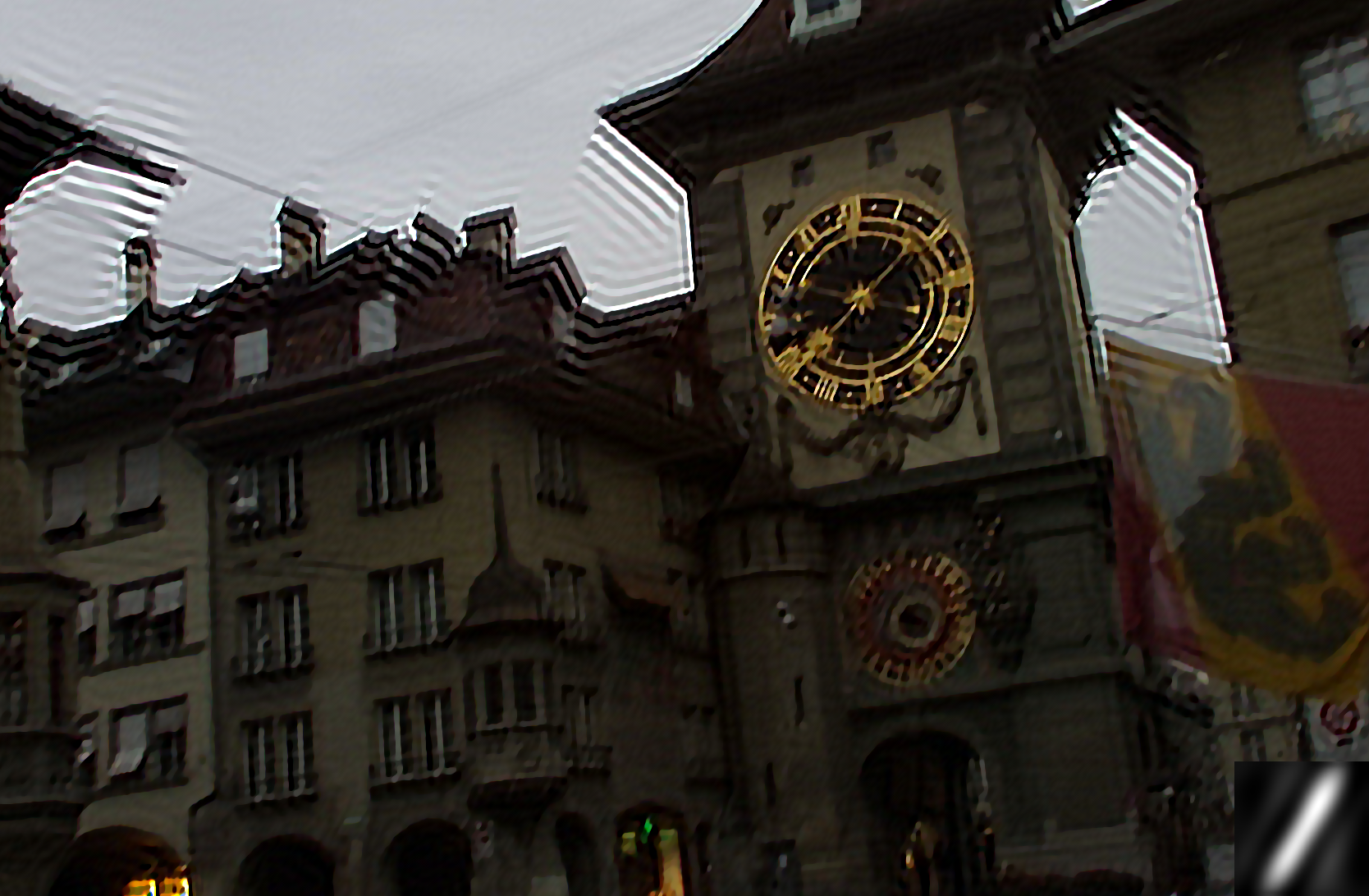}&\hspace{-12pt}
\includegraphics[width=167pt]{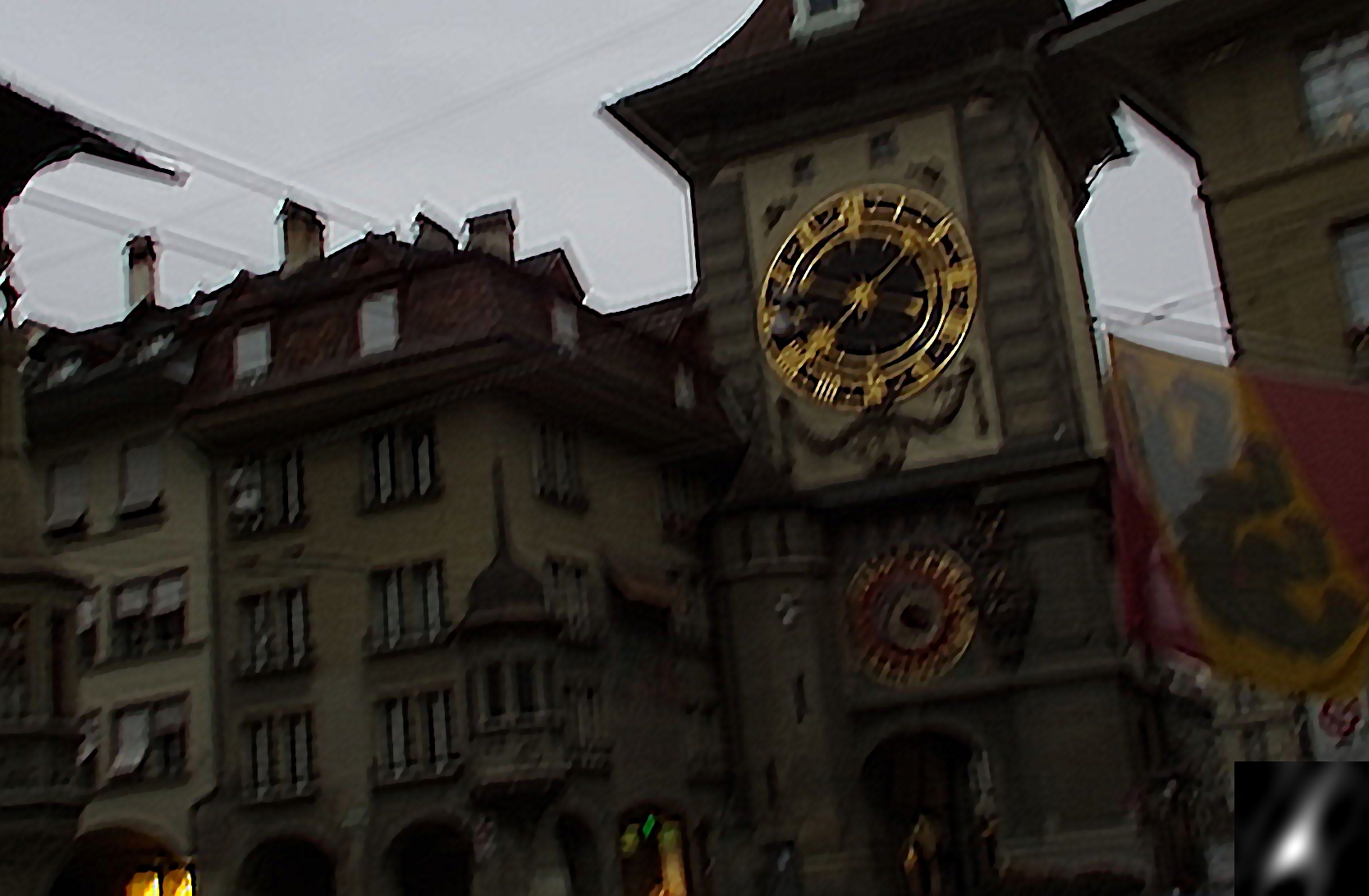}\\
\footnotesize{Output of \cite{ppr:whyte}}&\footnotesize{Deblurred using PSF estimated from \cite{babacan}}&\footnotesize{Deblurred using PSF estimated from \cite{logtv}}\\
\end{tabular}
\end{center}
\caption{Results on Zytglogge image.
\label{fig:Zy1}}
\end{figure*}

\subsection{Real experiments}
We perform real experiments using images of motion-blurred scenes captured with a Lytro Illum camera. For reading images from the Lytro Illum camera, we use the Light Field Toolbox V0.4 software \cite{web:toolb} developed by the authors of \cite{ppr:calib}. We initially apply intensity normalization to the raw image by using a white image of LF camera. The Lytro Illum camera consists of a database of white images for different camera settings. The Light Field Toolbox V0.4 has a function to pick the white image which is closest to a given image according to the camera settings. Intensity normalization helps to correct for the angular sensitivity of the image sensors. We captured motion blurred observations of three different real scenes from a handheld camera. In addition, we also captured another observation of the scene wherein the camera was placed on a support to avoid motion blur. We use this observation for visual evaluation of our result. In all our experiments, the scene was at a depth of more than three metres, and we used the same camera settings. We observed that for these parameters, if the depth value is beyond three meters, there would not be any variation in the LF PSF. This is because of the fact that in an LF camera the maximum observable baseline is limited to the lens aperture and consequently, beyond a certain depth the disparity becomes indistinguishable.

 The Lytro Illum camera has about $432{\times}540$ microlenses and the raw image has about $15$ pixels per microlens. For the LF image, we define a regular hexagonal grid with $Q=16$ and $Q^{\prime}=28$ (section \ref{sec:hex}) and calculate the affine warping matrix for this setting. The scene texture is defined on a grid with $B = 4$, and $B^{\prime} = 7$ which corresponds to about one-fourth the full sensor resolution. The LF PSF was obtained for this resolution of texture. We follow the approach discussed in section \ref{sec:practical} to estimate the motion PSF and thereby determine the sharp scene texture. For all experiments, the entire texture is divided into six overlapping patches in the form of a $2{\times}3$ array. i.e., we solve for six motion PSFs. In all our experiments, we used $\lambda = 1.6\cdot 10^{-3}$ for the alternating minimization scheme, and $\lambda = 4\cdot 10^{-4}$ for the final texture estimation. The motion PSF smoothness weight $\lambda_p$ was set to be $4000$.

  We compare the performance of the proposed method with \texttt{two-step} approach, implementation of our method with uniform motion blur assumption, and by applying the conventional blind deconvolution algorithms of \cite{ppr:whyte,babacan,logtv} on the refocused image rendered by Lytro software. In the \texttt{two-step} approach, we implemented a modified version of \cite{Perrone2014} wherein PSFs were estimated patch-wise. Similar to the proposed framework, in the \texttt{two-step} approach, we used six overlapping patches and imposed similarity constraints on the PSFs across patches.

\begin{figure*}[htb!]
\begin{center}
\begin{tabular}{ccc}
\includegraphics[width=167pt]{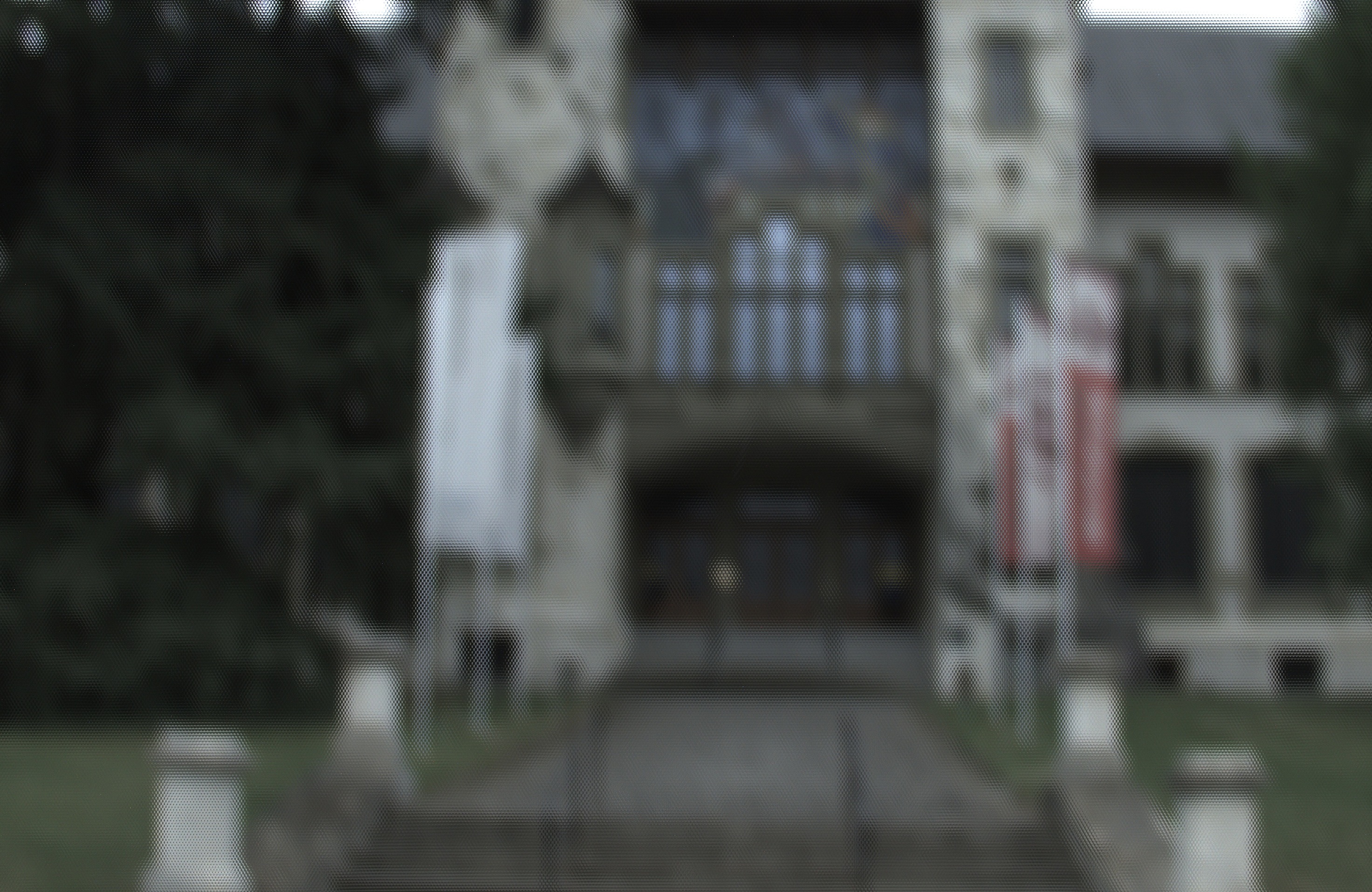}&\hspace{-12pt}
\includegraphics[width=167pt]{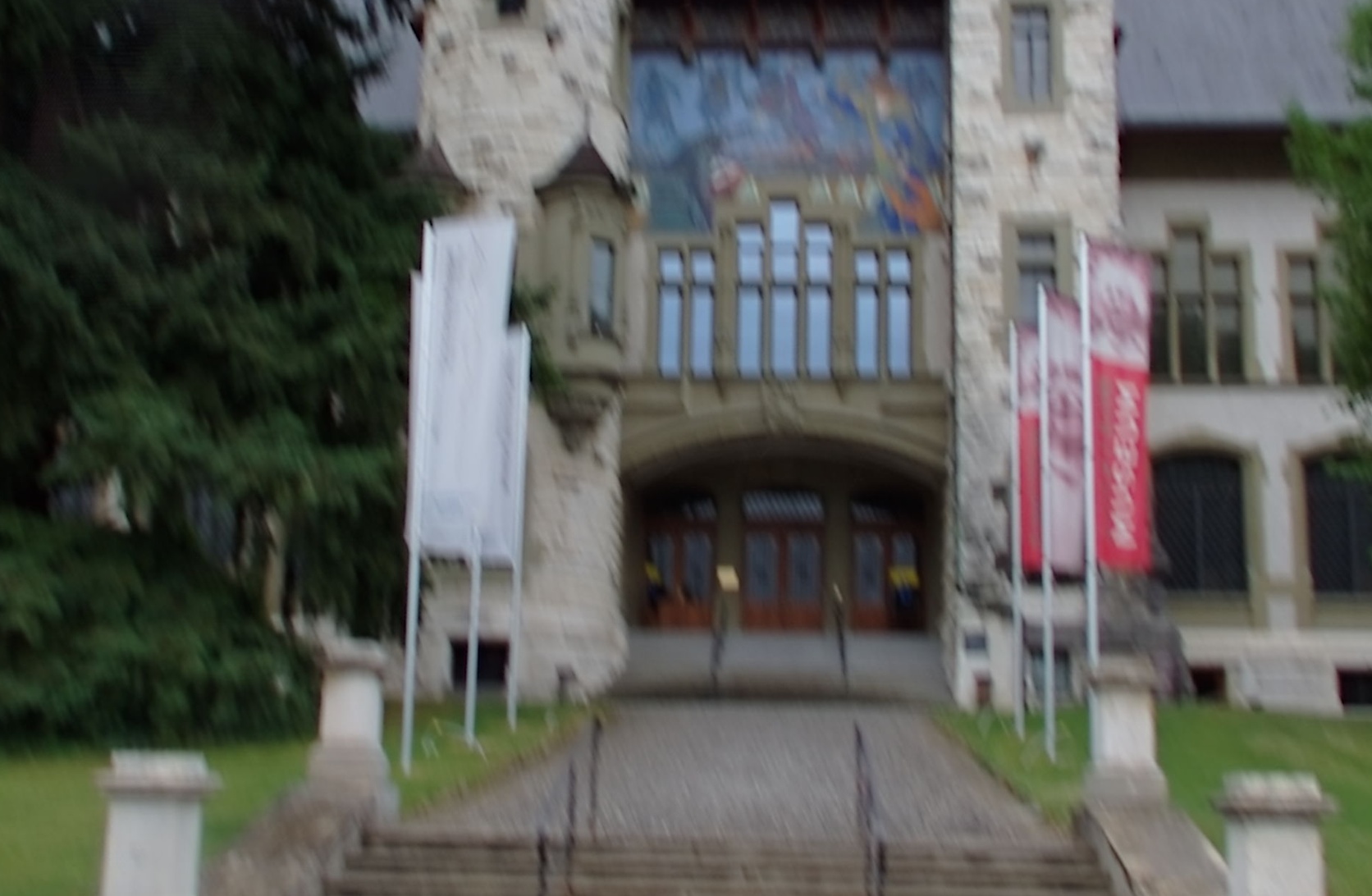}&\hspace{-12pt}
\includegraphics[width=167pt]{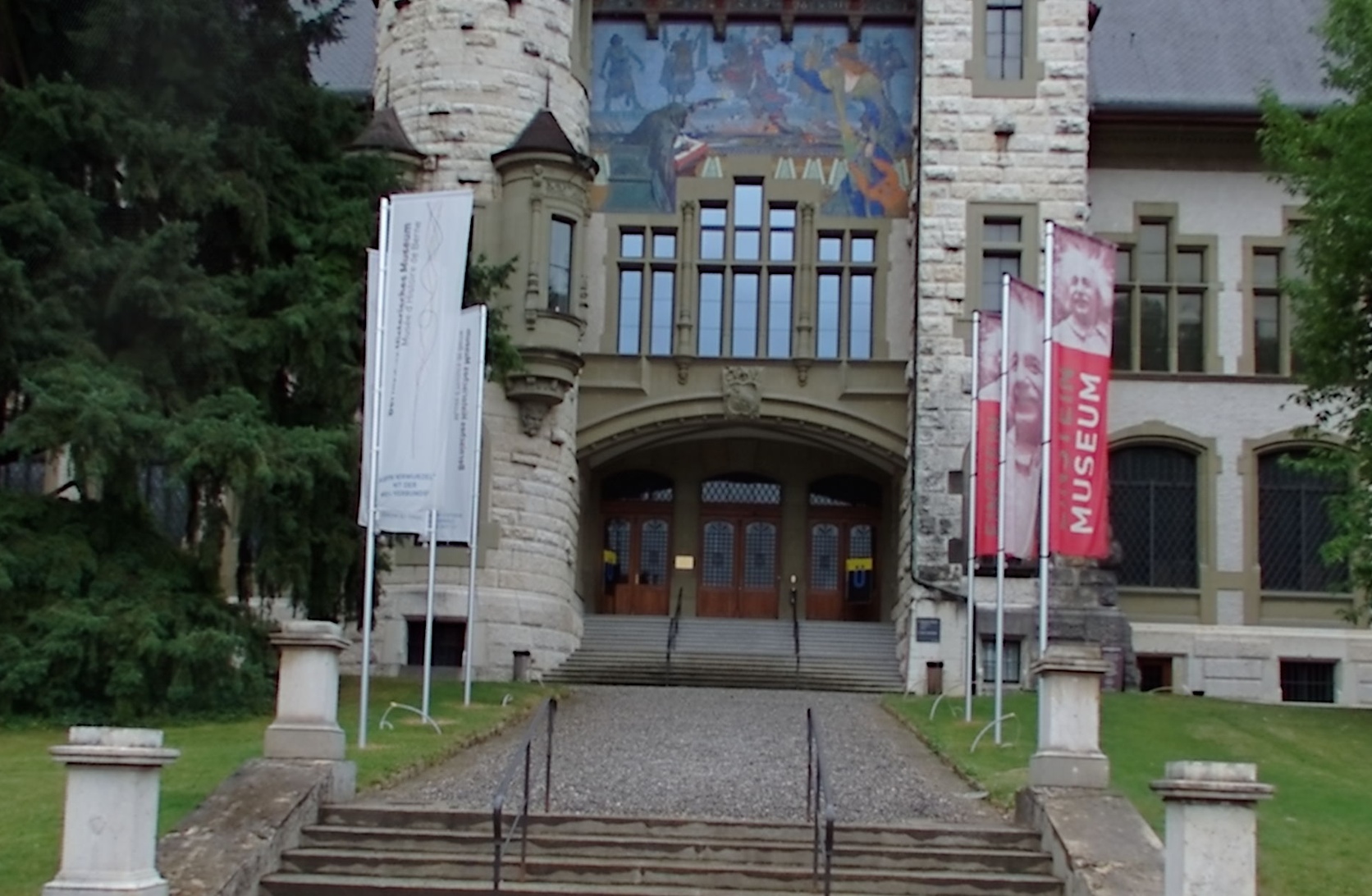}\\
\footnotesize{Raw LF image}&\footnotesize{Refocused image rendered by Lytro}&\footnotesize{Reference observation without motion blur}\\
\includegraphics[width=167pt]{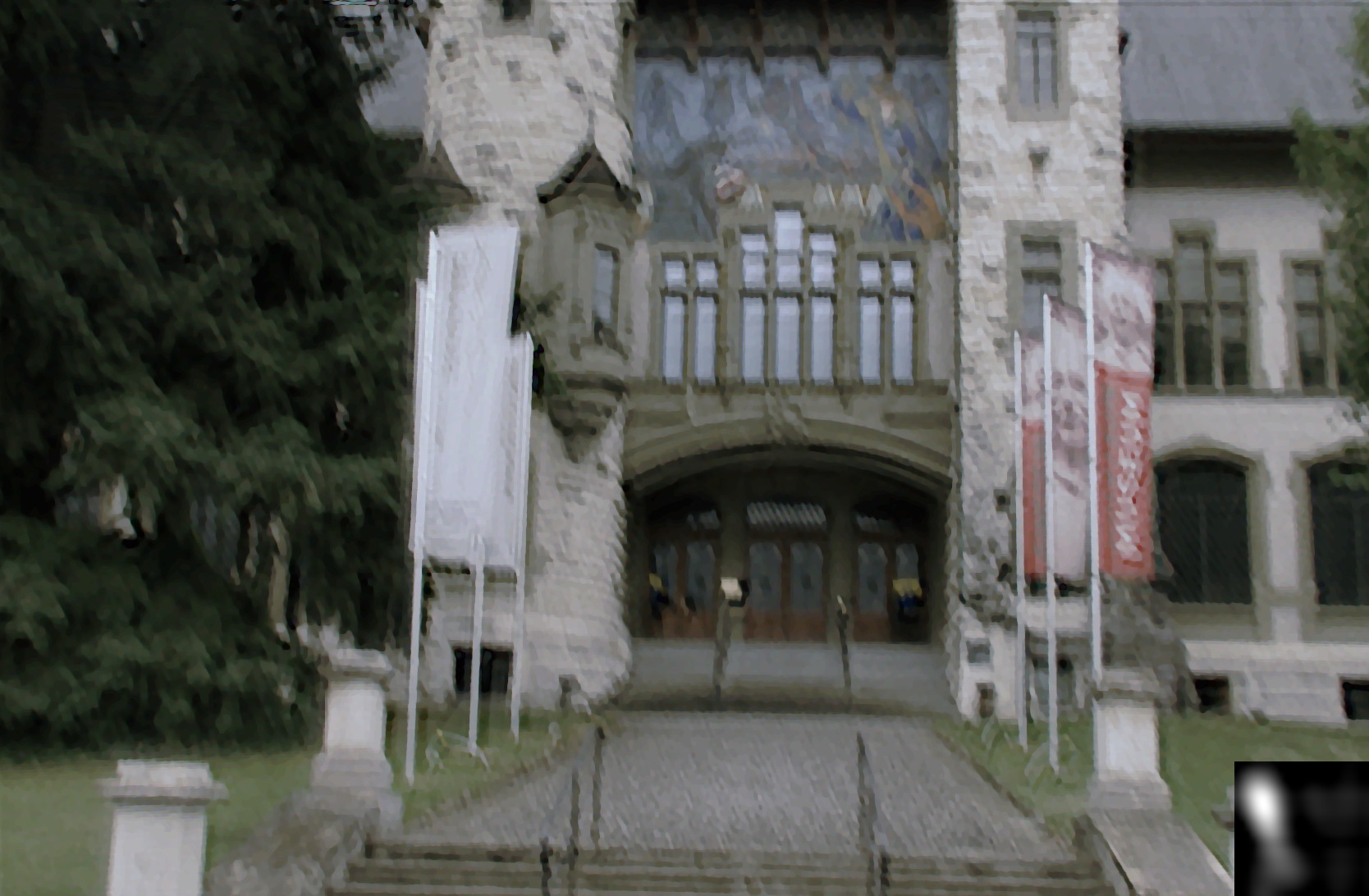}& \hspace{-12pt}
\includegraphics[width=167pt]{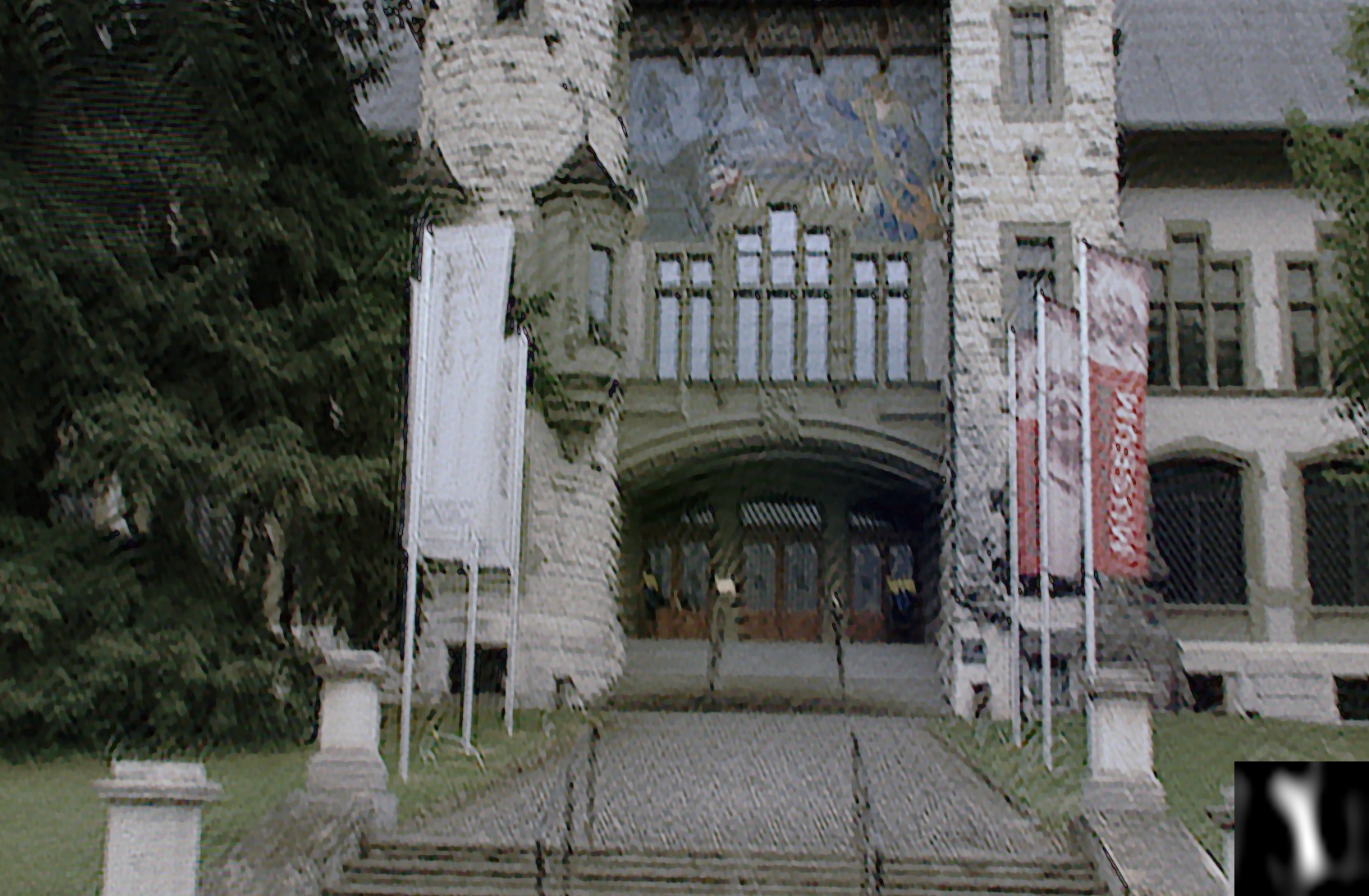}&\hspace{-12pt}
\includegraphics[width=167pt]{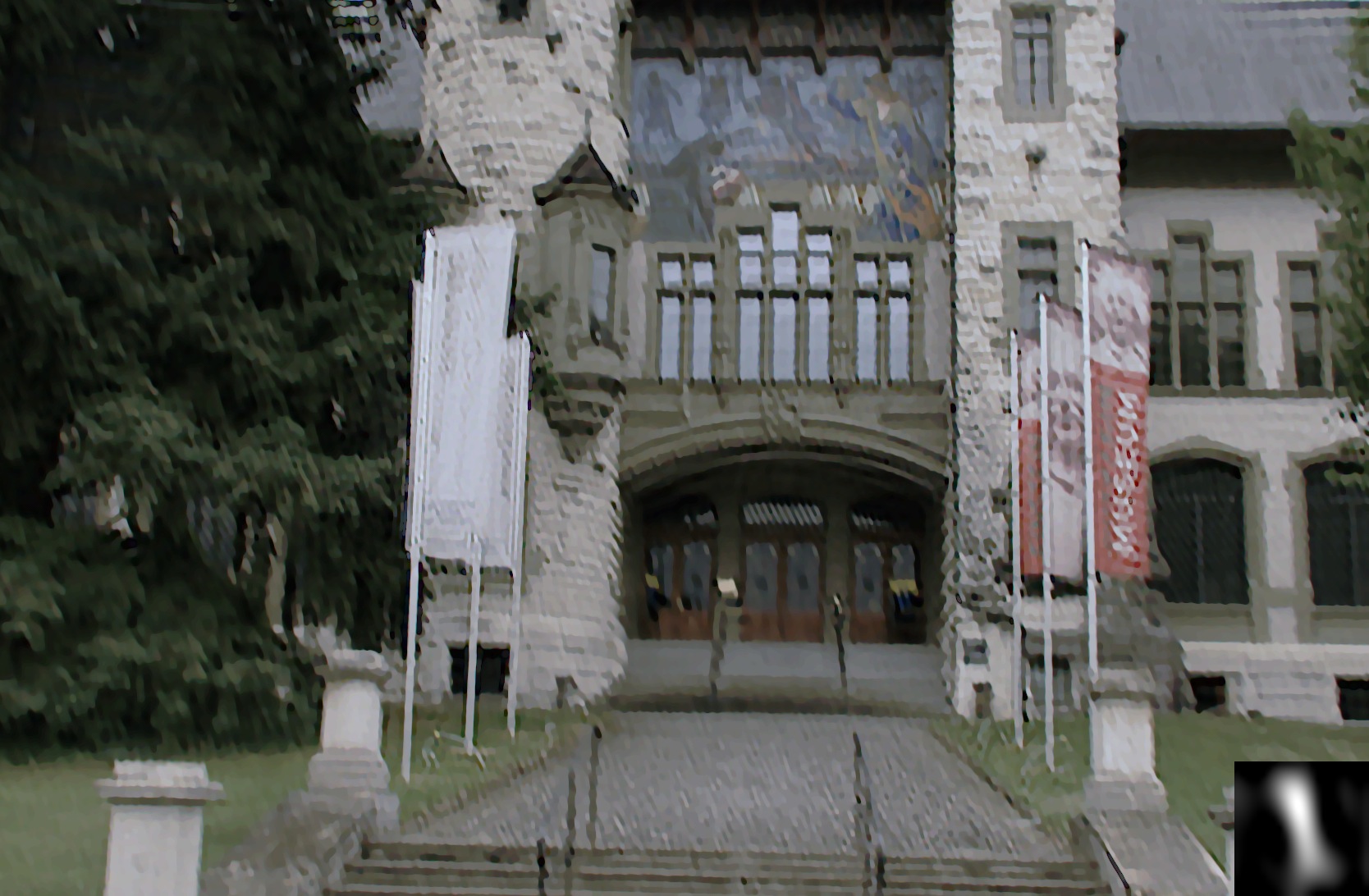}\\
\footnotesize{\texttt{two-step} output}&\footnotesize{Uniform motion blur approach}&\footnotesize{Proposed unified framework}\\
\includegraphics[width=167pt]{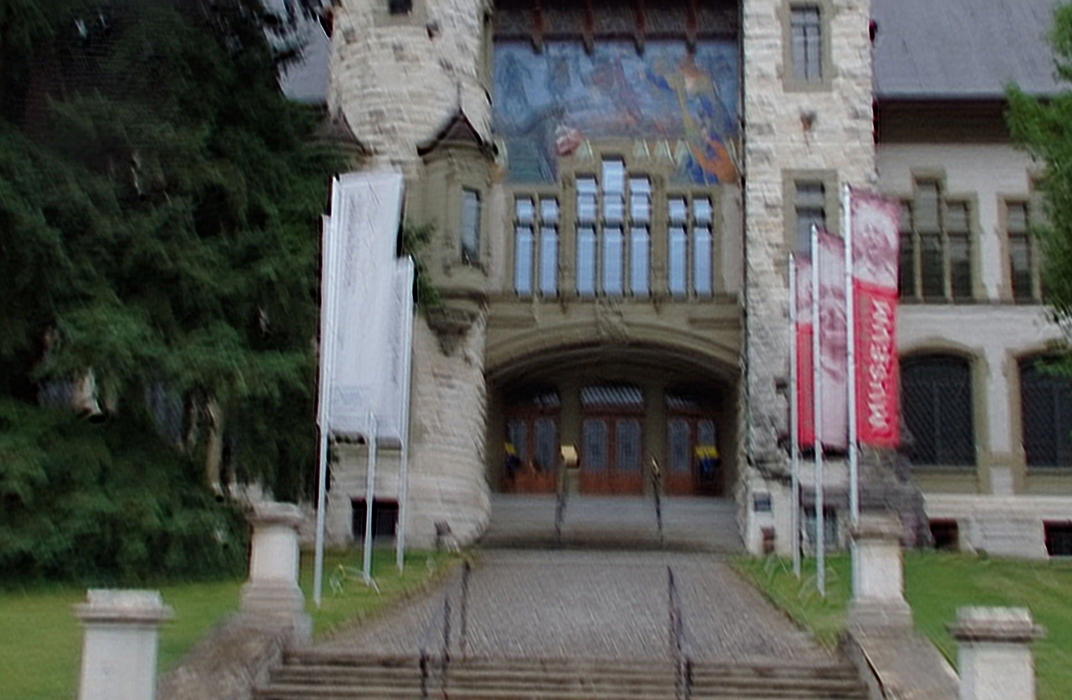}&\hspace{-12pt}
\includegraphics[width=167pt]{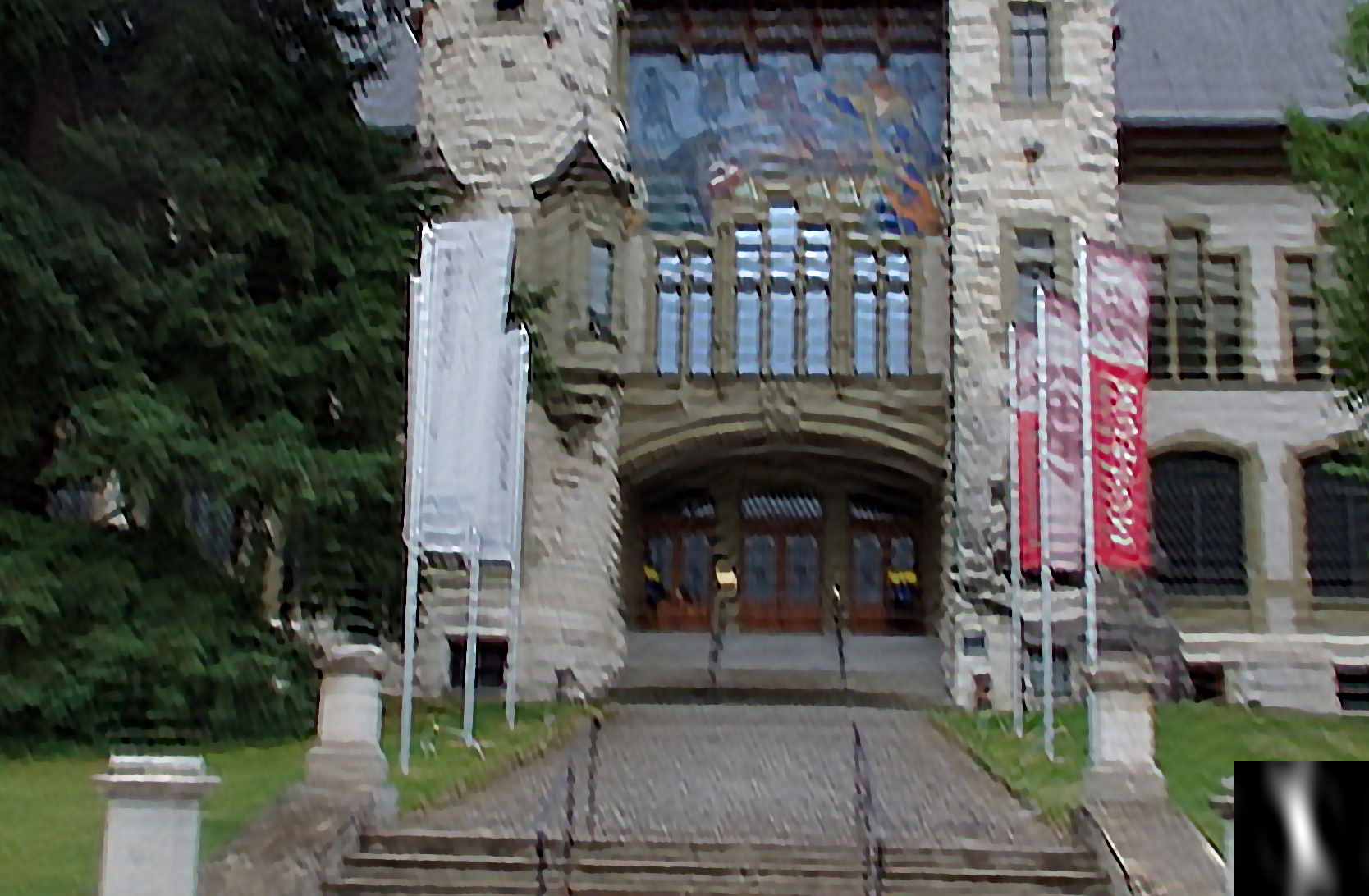}&\hspace{-12pt}
\includegraphics[width=167pt]{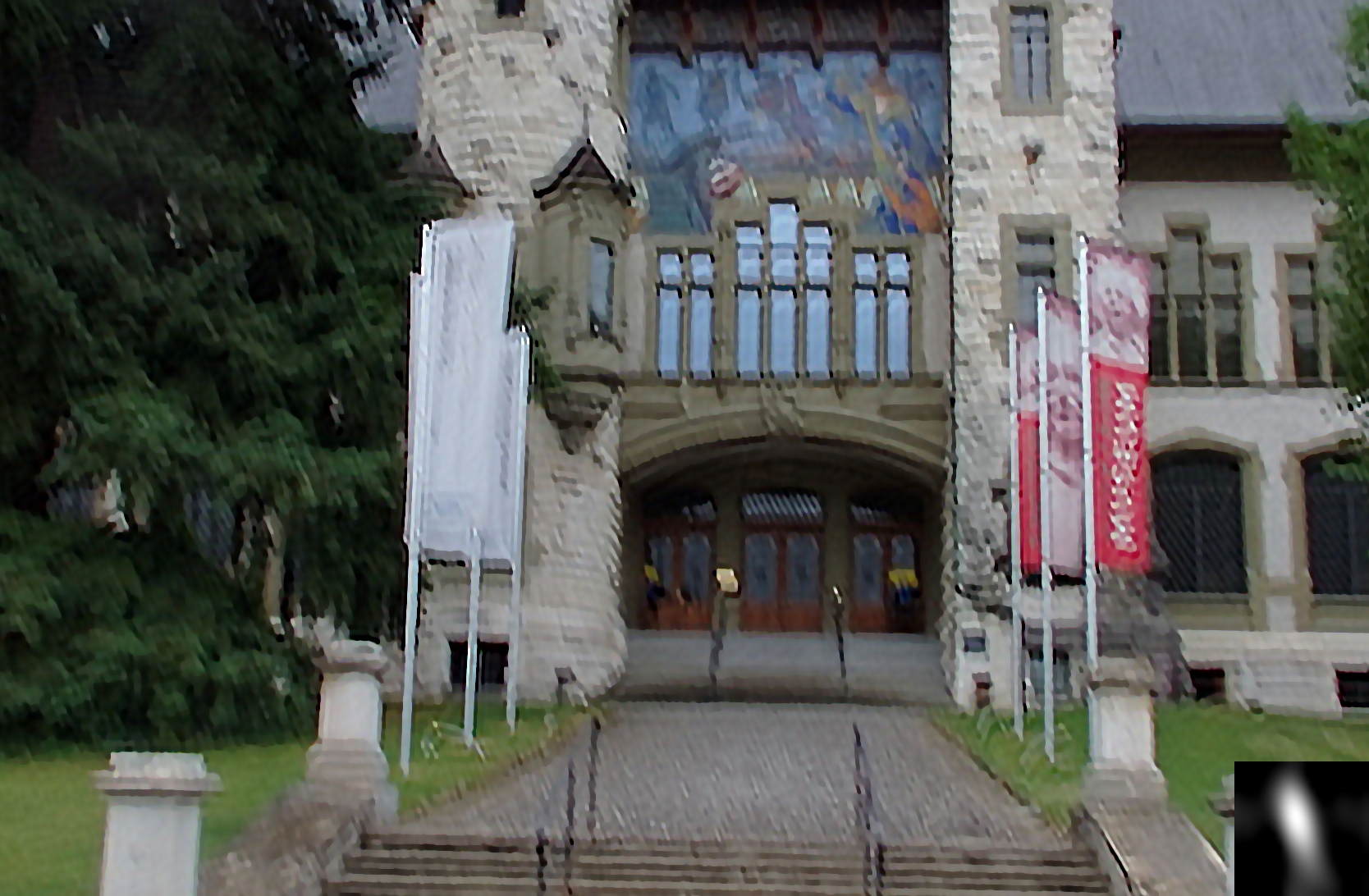}\\
\footnotesize{Output of \cite{ppr:whyte}}&\footnotesize{Deblurred using PSF estimated from \cite{babacan}}&\footnotesize{Deblurred using PSF estimated from \cite{logtv}}\\
\end{tabular}
\end{center}
\caption{Deblurring results on the image of a building.
\label{fig:H2}}
\end{figure*}

\begin{figure*}[htb!]
\begin{center}
\begin{tabular}{ccc}
\includegraphics[width=167pt]{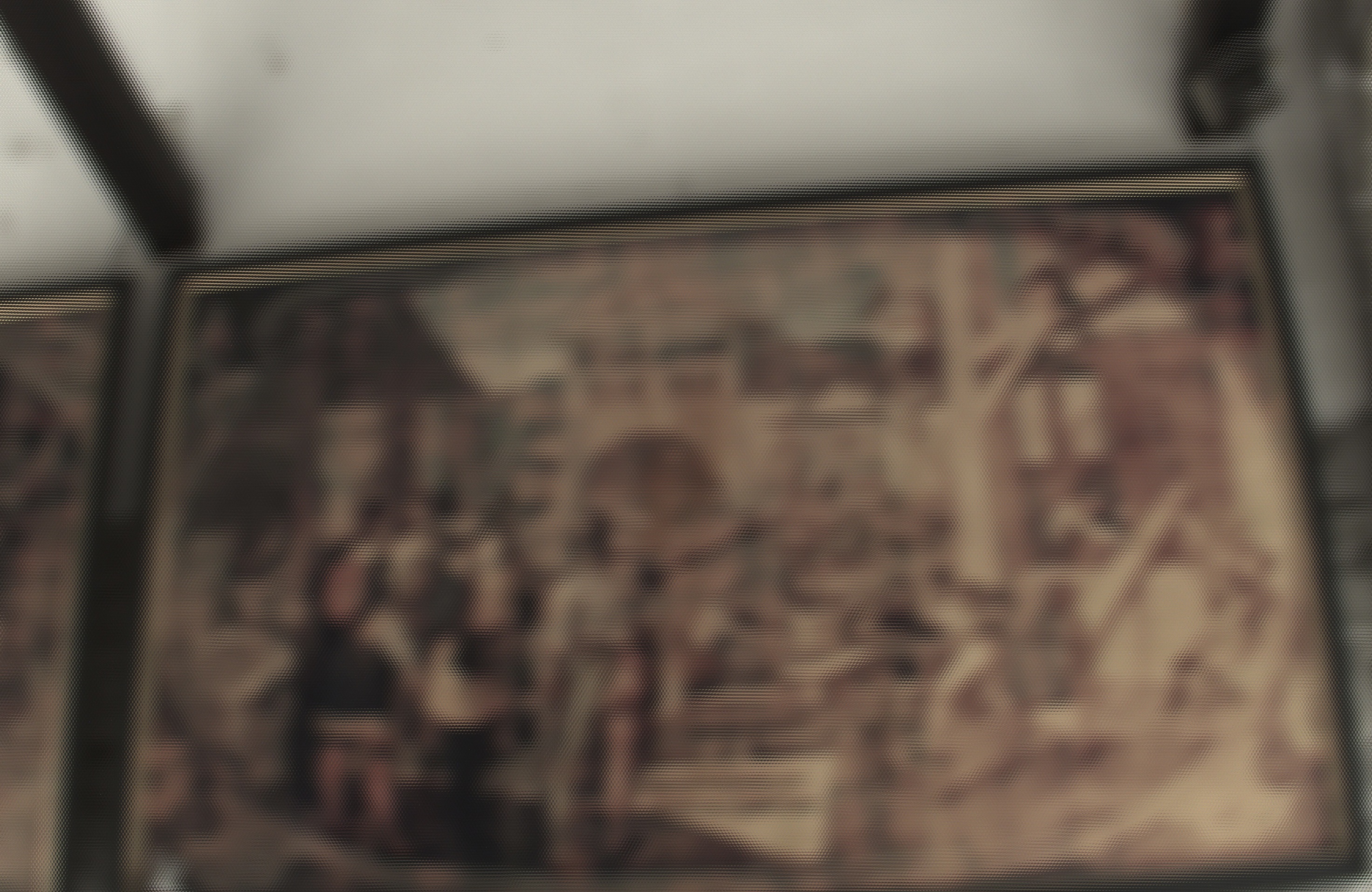}&\hspace{-12pt}
\includegraphics[width=167pt]{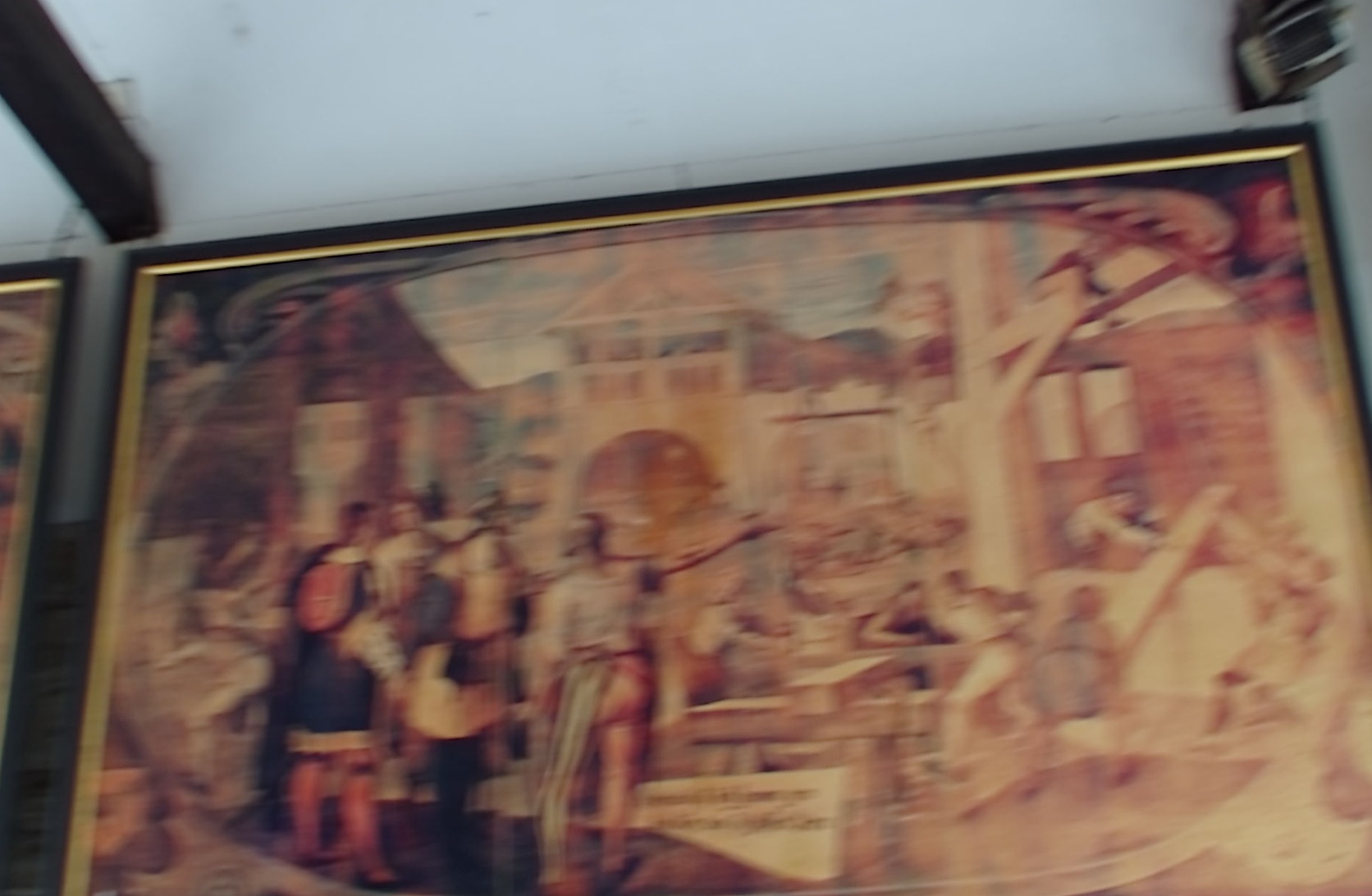}&\hspace{-12pt}
\includegraphics[width=167pt]{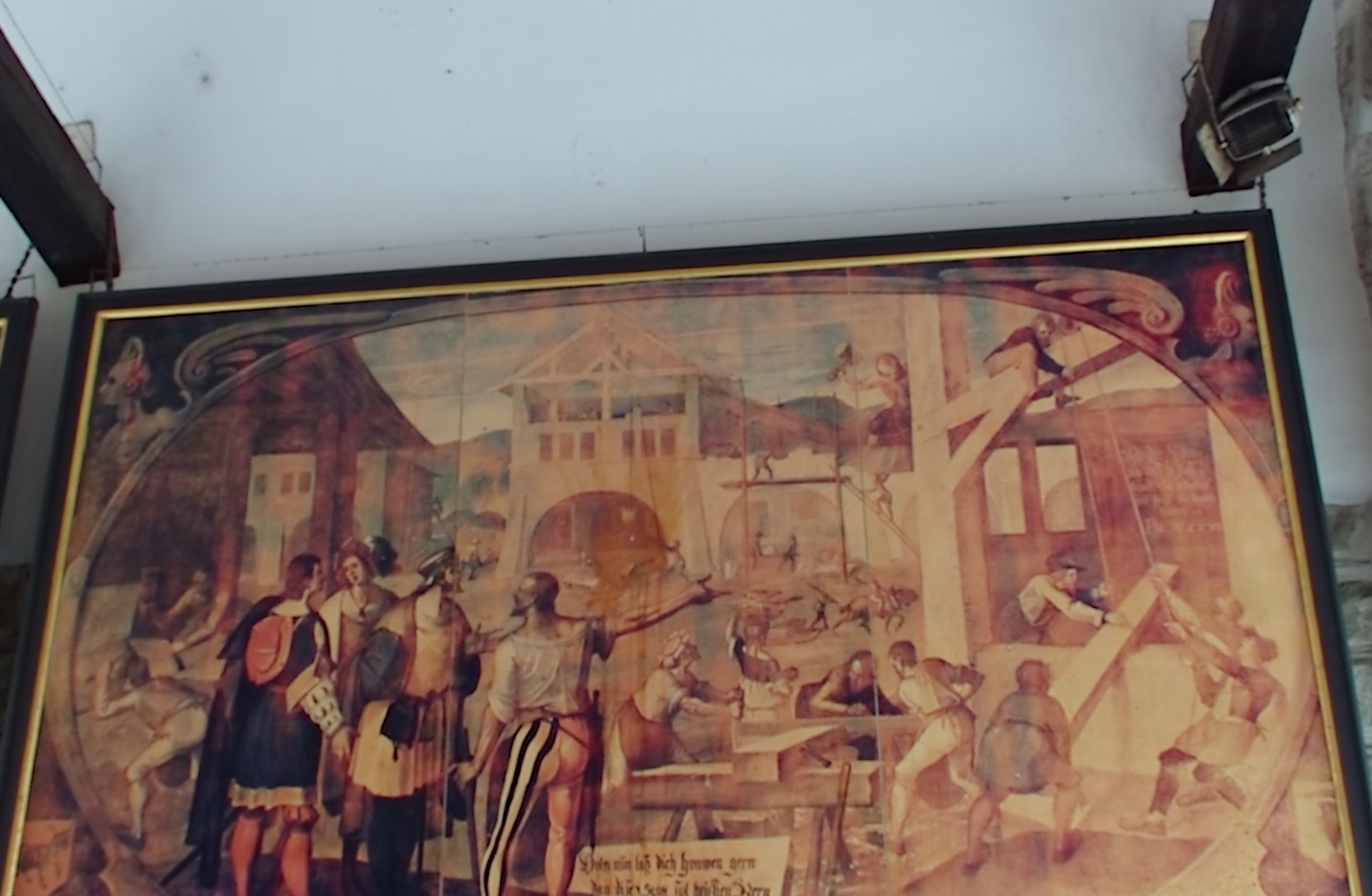}\\
\footnotesize{Raw LF image}&\footnotesize{Refocused image rendered by Lytro}&\footnotesize{Reference observation without motion blur}\\
\includegraphics[width=167pt]{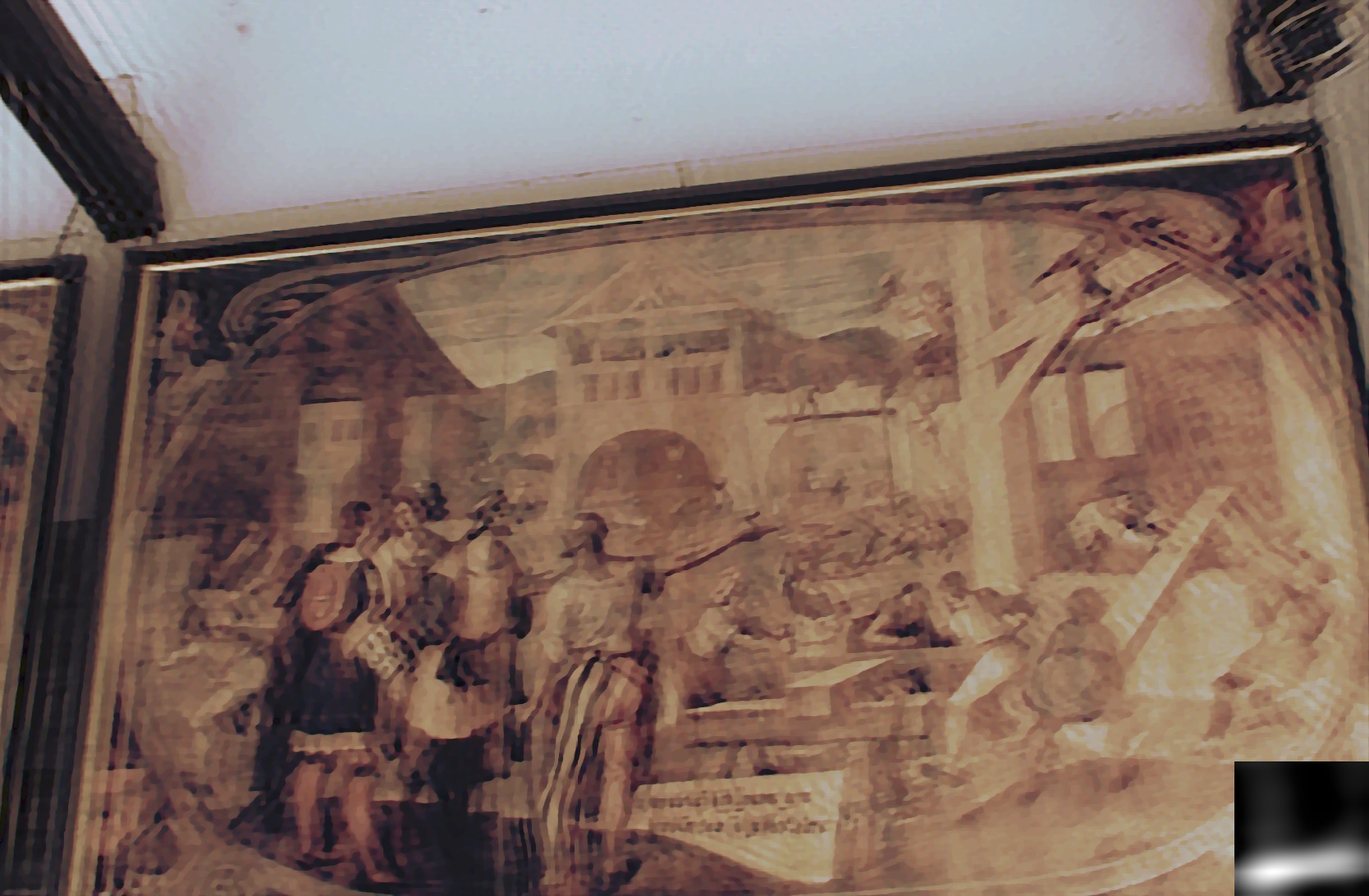}& \hspace{-12pt}
\includegraphics[width=167pt]{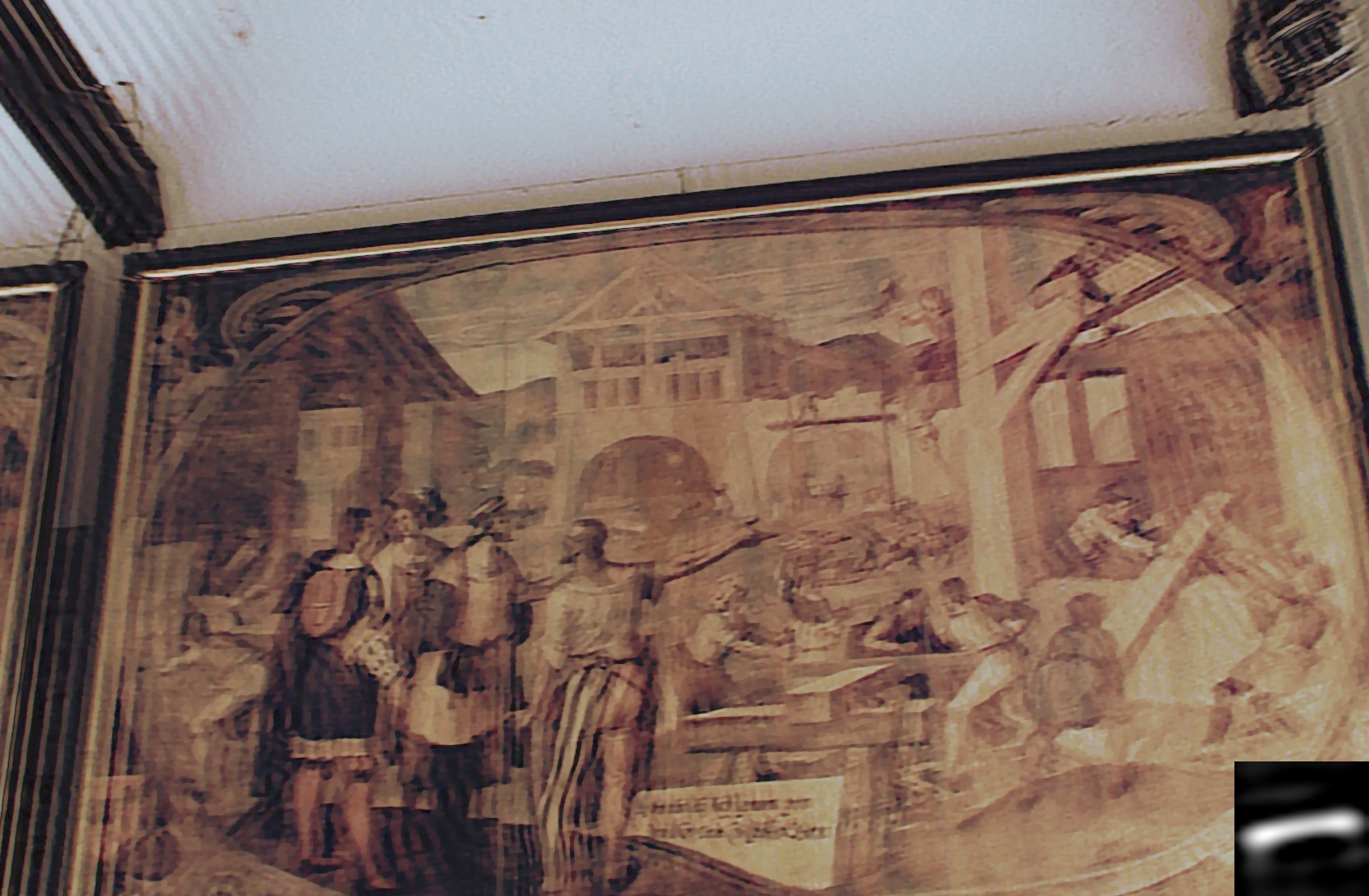}&\hspace{-12pt}
\includegraphics[width=167pt]{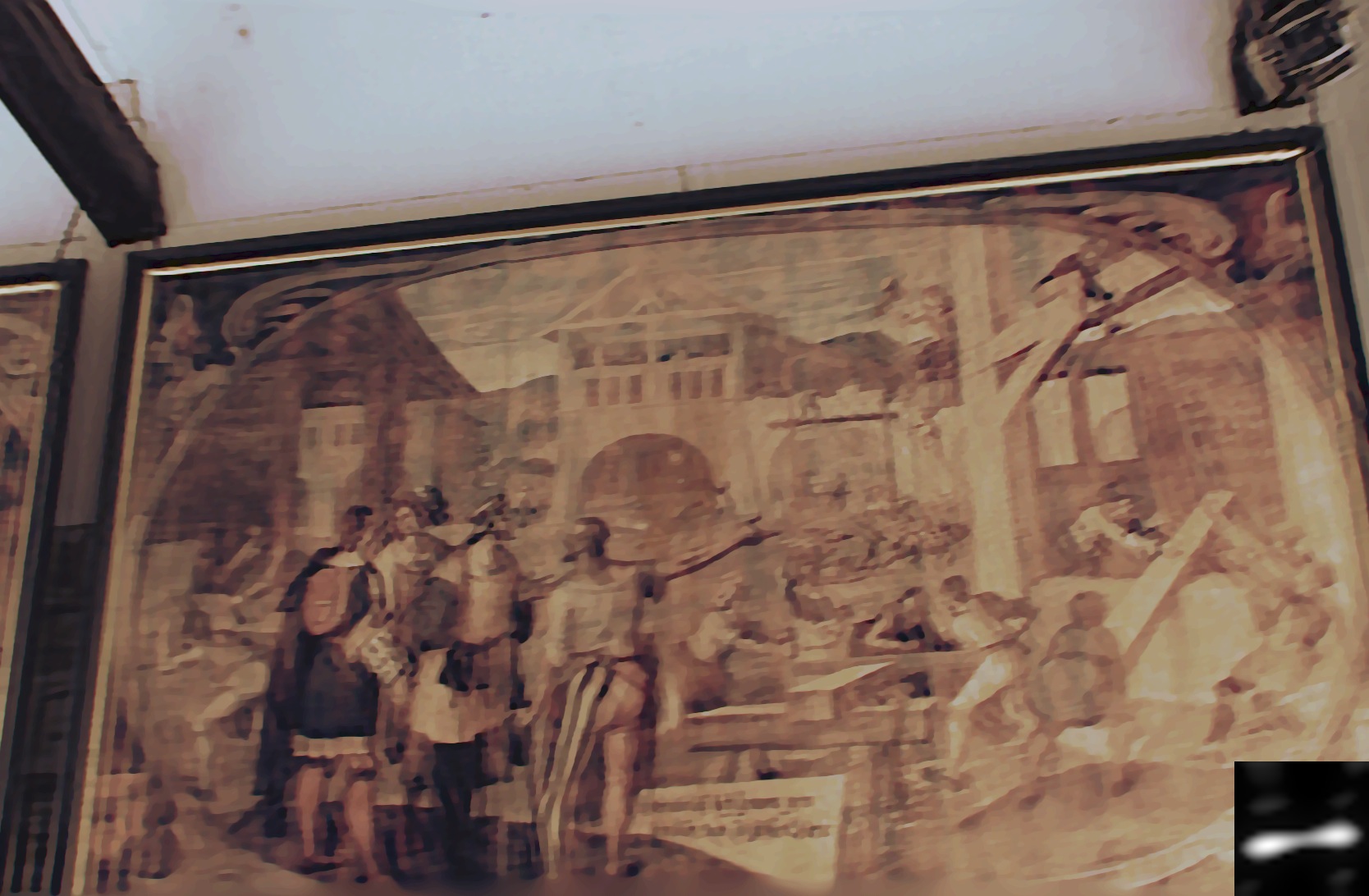}\\
\footnotesize{\texttt{two-step} output}&\footnotesize{Uniform motion blur approach}&\footnotesize{Proposed unified framework}\\
\includegraphics[width=167pt]{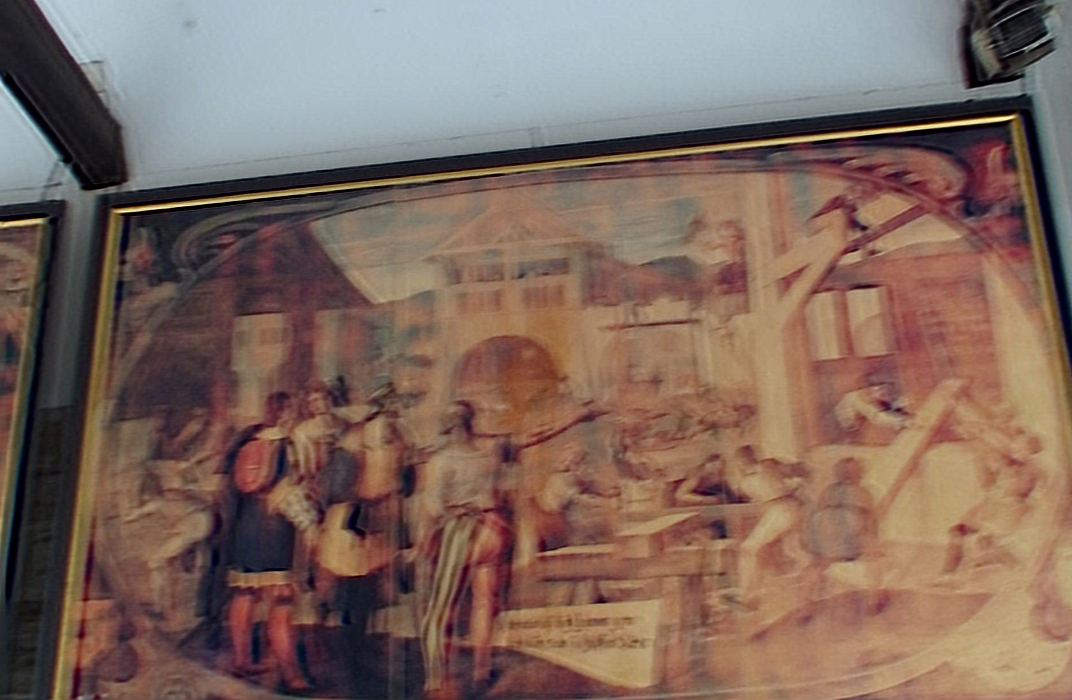}&\hspace{-12pt}
\includegraphics[width=167pt]{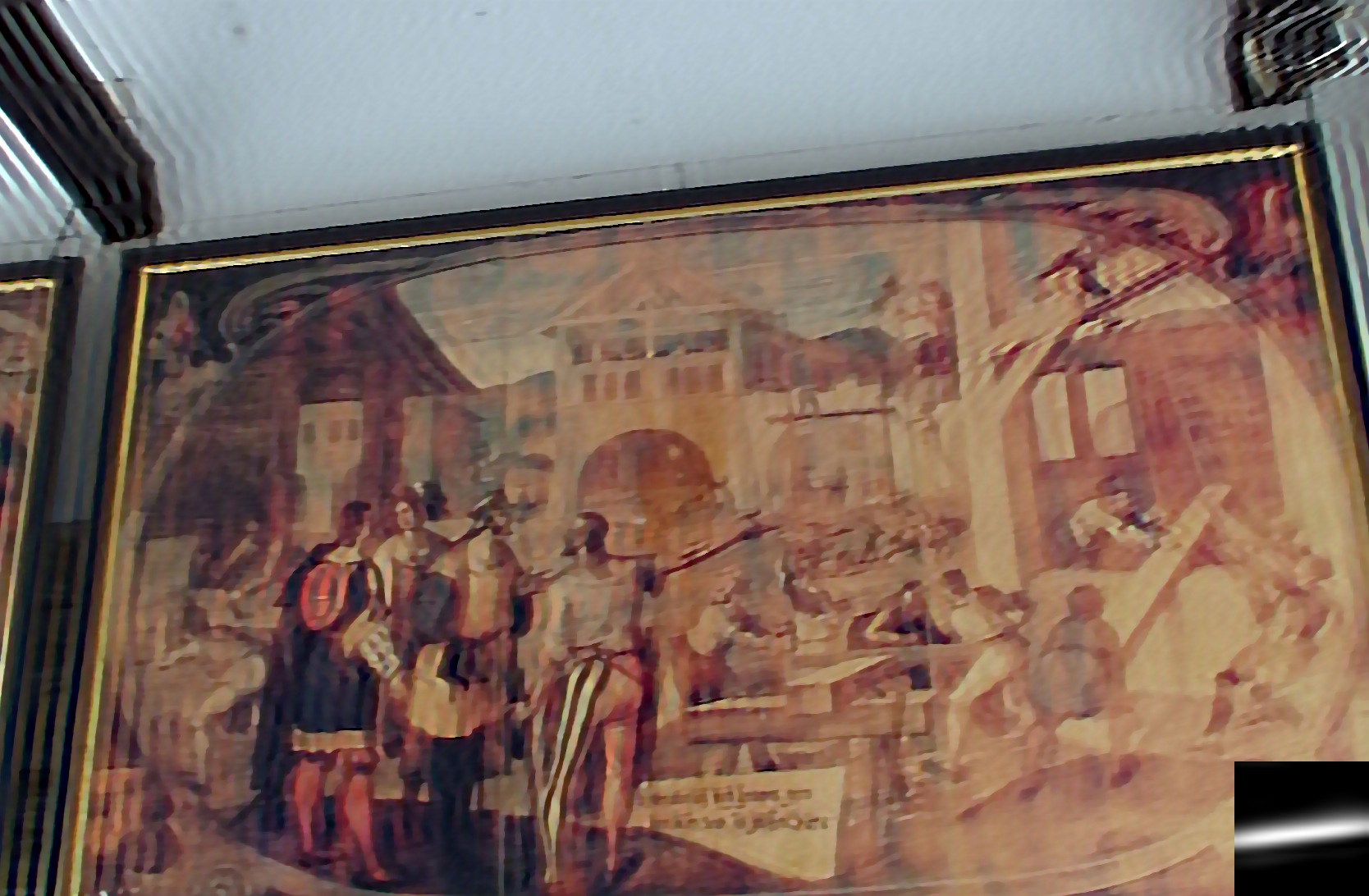}&\hspace{-12pt}
\includegraphics[width=167pt]{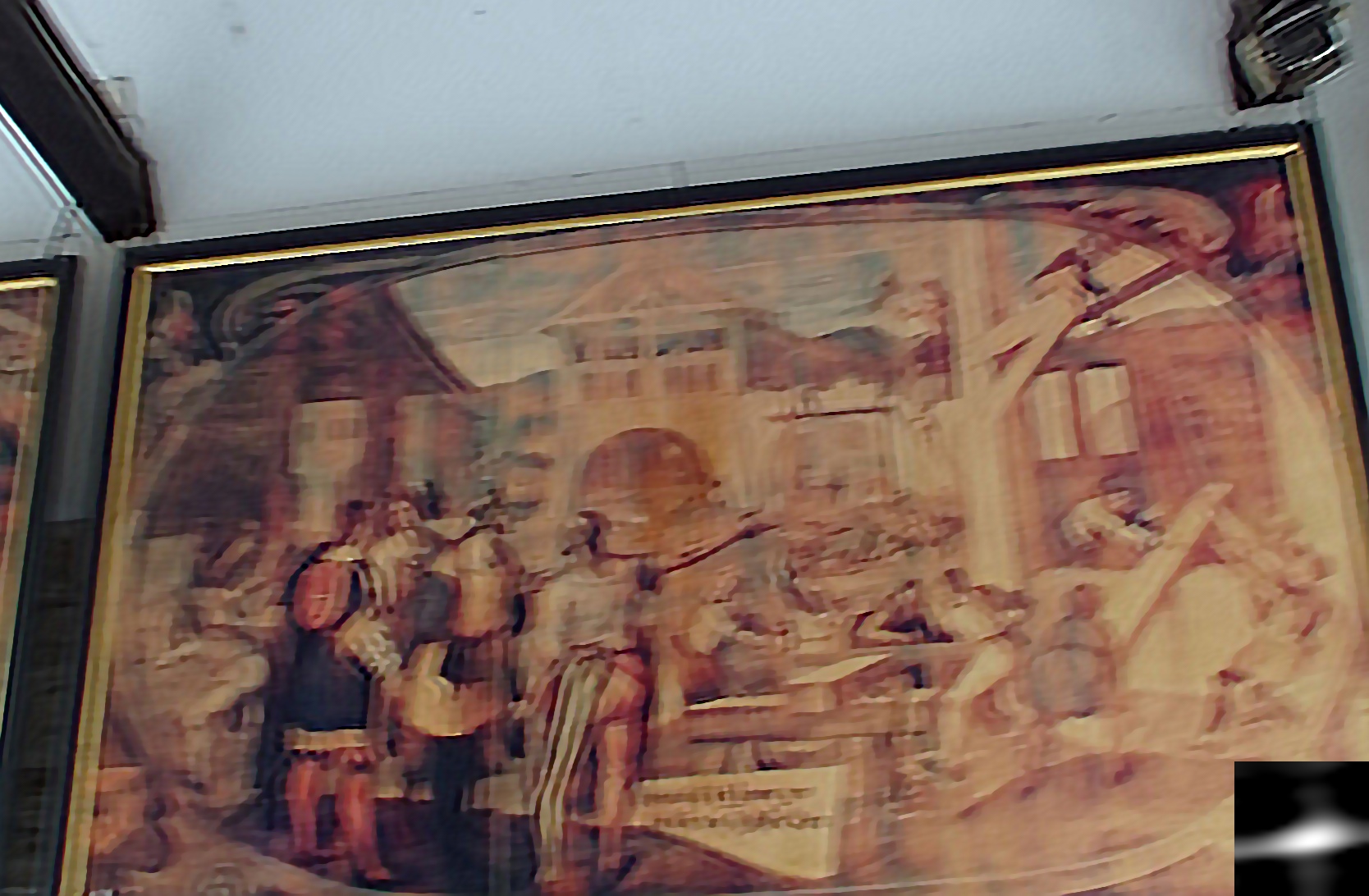}\\
\footnotesize{Output of \cite{ppr:whyte}}&\footnotesize{Deblurred using PSF estimated from \cite{babacan}}&\footnotesize{Deblurred using PSF estimated from \cite{logtv}}\\
\end{tabular}
\end{center}
\caption{Results on a scene consisting of a painting.
\label{fig:ZyPaint}}
\end{figure*}

We tested our algorithm on plenoptic images of four different scenes as shown in Figs. \ref{fig:post}-\ref{fig:ZyPaint}. In the first row of each of the figures, from left to right, we show raw image, refocused image rendered by Lytro, and rendering of another observation that was captured without any motion blur. In the second row, we show the output of \texttt{two-step} approach, result obtained by uniform motion blur assumption, and the result of the proposed scheme. We show only one motion PSF (at the bottom right inset) for the proposed result and the \texttt{two-step} approach instead of six PSFs because, there were only small changes in their values and visually they looked quite similar. In the third row we first show the result of applying the non-uniform deblurring algorithm in \cite{ppr:whyte} on the Lytro refocused image. The other two outputs in the third row were obtained by estimating motion blur PSFs from the Lytro rendered image using algorithms in \cite{babacan} and \cite{logtv}, followed by TV deblurring.

In all the examples, our approach significantly improves the quality of image when compared with the Lytro rendered image. For instance, in the Poster image (Fig. \ref{fig:post}) not only the text on the posters but also the fine texture on the background wall is similar to the texture in the unblurred observation. In the Painting image (Fig. \ref{fig:ZyPaint}), the features are sharp across the entire image in the proposed scheme as compared to other outputs. The reader is encouraged to zoom on all the observations for better examination. Although the outputs from other techniques do exhibit deblurring at instances, they also contain ringing artifacts and residual blur.

\section{Conclusions}
We have presented the first motion deblurring method for plenoptic cameras. Our method extends classical blind deconvolution methods to light field cameras by modeling the interaction of motion blur and plenotic point spread functions. Moreover, our model is highly parallelizable and memory efficient. We follow an alternating minimization procedure to estimate the sharp and super resolved texture and the unknown motion blur. The practical issues of alignment and radial distortion are accounted for in our energy minimization framework. Experiments show that the performance of our method is quite good and outperforms approaches based on conventional blind deconvolution. Our method can be considered as a stepping stone towards more generalized scenarios of 3D scenes and non-uniform motion blur.

\end{document}